\documentclass[letterpaper, 10 pt, conference]{ieeeconf}  
\IEEEoverridecommandlockouts                              
\usepackage{amsmath,amsfonts,amssymb}
\usepackage{algorithmic}
\usepackage{algorithm}
\usepackage{xcolor}
\usepackage{breqn}
\usepackage{array}
\usepackage{textcomp}
\usepackage{stfloats}
\usepackage{url}
\usepackage{verbatim}
\usepackage{graphicx}
\usepackage{cite}
\usepackage{multirow}
\usepackage{mathtools}
\usepackage{subcaption}

\usepackage{ragged2e}
\usepackage{balance}

\title{\LARGE \bf
    Safe CoR: A Dual-Expert Approach to Integrating Imitation Learning and Safe Reinforcement Learning Using Constraint Rewards
}

\author{Hyeokjin Kwon$^1$, Gunmin Lee$^2$, Junseo Lee$^1$, Songhwai Oh$^{1,2}$
\thanks{$^{1}$ H. Kwon, J. Lee, and S. Oh are with the Interdisciplinary Program in Artificial Intelligence and ASRI, Seoul National University, Seoul 08826, Korea (e-mail: hyeokin.kwon@rllab.snu.ac.kr, junseo.lee@rllab.snu.ac.kr, songhwai@snu.ac.kr)
$^{2}$ G. Lee and S. Oh are with the Department of Electrical and Computer Engineering and ASRI, Seoul National University, Seoul 08826, Korea (e-mail: gunmin.lee@rllab.snu.ac.kr).}
}

\begin{document}
\maketitle
\thispagestyle{empty}
\pagestyle{empty}

\begin{abstract}

In the realm of autonomous agents, ensuring safety and reliability in complex and dynamic environments remains a paramount challenge. Safe reinforcement learning addresses these concerns by introducing safety constraints, but still faces challenges in navigating intricate environments such as complex driving situations. To overcome these challenges, we present the safe constraint reward (Safe CoR) framework, a novel method that utilizes two types of expert demonstrations—reward expert demonstrations focusing on performance optimization and safe expert demonstrations prioritizing safety. By exploiting a constraint reward (CoR), our framework guides the agent to balance performance goals of reward sum with safety constraints. We test the proposed framework in diverse environments, including the safety gym, metadrive, and the real-world Jackal platform. Our proposed framework enhances the performance of algorithms by 39\% and reduces constraint violations by 88\% on the real-world Jackal platform, demonstrating the framework's efficacy. Through this innovative approach, we expect significant advancements in real-world performance, leading to transformative effects in the realm of safe and reliable autonomous agents.

\end{abstract}

\section{Introduction}
The advance of autonomous driving technology promises to revolutionize the way people commute, offering safer, more efficient, and accessible transportation options. At the heart of this transformative potential is the importance of ensuring the safety and reliability of autonomous vehicles in diverse and dynamic driving environments. To achieve this, many researchers and engineers have proposed algorithms such as rule-based controllers\cite{rb1, rb2} and imitation learning methods \cite{transfuser, MixGAIL, ssil}. Rule-based controllers provide a structured approach to decision-making based on predefined rules and conditions, while imitation learning allows the agents to mimic human driving behaviors by learning from vast amounts of driving data. However, these methods face significant challenges in handling situations that fall beyond predefined rules \cite{intro1}. These scenarios, which are neither encapsulated within the training data nor foreseen in the predefined rule sets, pose critical hurdles in achieving the comprehensive coverage and reliability that autonomous driving aspires to achieve.

To address the limitations inherent in imitation learning and rule-based controllers, reinforcement learning (RL) \cite{rl1, rl2} has emerged as a compelling alternative. Unlike its predecessors, RL enables autonomous driving agents to learn optimal behaviors through trial and error, interacting directly with their environment. This method offers significant advantages, such as the ability to continuously improve and adapt to new situations over time, potentially covering the gaps left by imitation learning and rule-based systems. Although RL excels in adaptability and decision-making in complex scenarios, ensuring the safety of autonomous driving agents remains a critical challenge. However, the exploratory nature of RL, which often requires agents to make mistakes to learn, poses a significant risk in real-world driving contexts where safety is crucial. This fundamental concern highlights the need for innovative approaches within RL frameworks to balance exploration with the stringent safety requirements of autonomous driving.

To address the aforementioned issue, the concept of safe reinforcement learning (safe RL) \cite{CPO, TRC} has been introduced. This approach aims to incorporate safety constraints into the optimization process explicitly. By taking account of safety constraints into the policy optimization process, safe RL methods enhance the agent's ability to adhere to safety constraints, thereby improving safety during both the training phase and the final deployment. For instance, incorporating a lane-keeping reward directly into the reward function results in mediocre lane-keeping behavior. On the other hand, when the lane-keeping component is applied as a constraint within the safe RL framework, the agent demonstrates significantly improved lane-keeping performance. Despite these advancements, challenges persist in the application of safe RL algorithms for training agents to navigate complex driving environments safely.

To overcome these challenges, we propose a novel method called safe CoR, which innovatively combines two distinct types of expert demonstrations to refine existing safe RL algorithms. The first type, termed reward expert demonstrations, focuses exclusively on maximizing rewards without considering safety constraints. Conversely, the second type, safe expert demonstrations, prioritizes adherence to safety requirements above all, with subsequent consideration for reward maximization. By distinctly categorizing these experts—reward experts for their focus on performance optimization and safe experts for their dual focus on safety and reward maximization—we are able to calculate a constraint reward (CoR). This term aids in the update process, directing the agent to emulate the reward expert for maximizing rewards while using the safe expert as a regularizer to ensure constraint satisfaction. Through the strategic application of CoR, our method guides the agent toward reducing constraint violations (CV) while still achieving high levels of reward, illustrating a balanced approach to learning optimal behaviors in diverse driving conditions. This dual-expert framework significantly enhances the agent's ability to navigate complex driving scenarios, striking a critical balance between ambitious performance goals and stringent safety standards.

Our experimental outcomes demonstrate that the safe CoR framework significantly improves algorithmic performance while diminishing constraint violations across various platforms, including the metadrive simulator \cite{metadrive} and safety gym environments \cite{safetygym}. Notably, when applied to the real-world Jackal platform \cite{TRC}, our framework achieved superior results over simulated environments, empirically demonstrating the advantage of the proposed framework. These findings underscore safe CoR's substantial potential in advancing the domain of safe RL.

The contributions of this paper are summarized as follows:
\begin{itemize}
    \item We propose a framework called safe CoR, which uniquely integrates reward-centric and safety-conscious expert data to refine and enhance the performance of existing safe RL algorithms in the autonomous driving domain.
    \item We show empirical evidence demonstrating that agents, under the guidance of the safe CoR framework, outperform traditional safe RL algorithms by achieving superior performance metrics, especially in the real-world platform, with reduced rates of constraint violations in the training phase.
    \item We validate the superiority of the proposed algorithm in real-world scenarios utilizing the Jackal robot platform, thereby affirming the framework's applicability and robustness across diverse operational environments.
\end{itemize}

\section{Related Work}
\subsection{Imitation learning}
Imitation learning is one of the main approaches in achieving autonomous driving agents. It is a method that guides agents to imitate the given demonstrations extracted from experts. One of the simplest approaches to imitation learning is behavior cloning (BC), which shows promising results in achieving generalization in real-world environments \cite{bc, bc2}. Despite its promise, BC is particularly susceptible to compounding errors, a drawback that significantly hampers its effectiveness \cite{bcerror}. On the other hand, inverse reinforcement learning (IRL) \cite{IRL} proposes another way to solve the problem of designing an autonomous agent, which is to learn the reward function from the expert demonstrations. Ho et al. \cite{gail} proposed an algorithm that integrates IRL and RL, enabling the agent to acquire expert behaviors and estimate the reward function concurrently. They mathematically proved the convergence of training both policies and discriminators alternatively and their research opened avenues for further researchers \cite{MixGAIL, infogail, wassersteingail}.

Additionally, there have been studies that combine imitation learning with online learning. Yiren et al. \cite{bcsac} experimentally demonstrated that expert demonstrations can assist agents in navigating challenging environments robustly. Despite these advancements, it is crucial to note that the mentioned methods have limitations as they do not directly account for safety constraints in the learning process.

\subsection{Safe reinforcement learning}
Safe reinforcement learning (safe RL) addresses the critical aspect of satisfying the safety of agents by integrating safety considerations into the learning process. The algorithm forces agents not only to maximize reward sums but also to satisfy given constraints simultaneously. This approach can be categorized into two methods: Lagrangian-based and trust-region-based methods. 

Lagrangian-based method transforms the original safe RL problem into its dual problem. Ray et al. \cite{safetygym} proposed the proximal policy optimization-Lagrangian (PPO-Lagrangian) algorithm, which extends the traditional PPO framework by incorporating a Lagrangian multiplier approach to efficiently handle constraints, allowing for dynamic adjustment of the trade-off between policy performance and constraint satisfaction. Yang et al. \cite{WCSAC} proposed the worst-case soft actor-critic (WCSAC), which relaxes constrained problems to unconstrained ones using Lagrangian multipliers. However, such algorithms suffer from being overly conservative in their updates when constraint violations occur excessively during the initial learning stage. Additionally, the usage of Lagrangian multipliers makes the learning process unstable.

Trust-region-based method is an extended version of trust region policy optimization \cite{TRPO}, which solves non-convex optimization by transforming it into a simple problem. Achiam et al. \cite{CPO} introduced constrained policy optimization (CPO), which addresses the optimization of policy functions under safety constraints without transforming them into different forms of optimization problems. CPO uniquely maintains safety constraints by utilizing a trust region method, ensuring that policy updates remain within predefined safety limits, thereby facilitating the development of safe reinforcement learning policies. Kim and Oh proposed TRC and OffTRC \cite{TRC, OffTRC}, assuming that the discounted cost sum follows a Gaussian distribution. They derived the closed-form upper bound of conditional value at risk (CVaR). Recently, Kim et al. \cite{SDAC} proposed a method that utilizes a distributional critic and gradient-integration technique to enhance the stability of the agent. However, the above algorithms still face challenges in learning agents for safe driving in complex environments.

\section{Preliminary}
\subsection{Constrained Markov decision process}
A constrained Markov decision process (CMDP) is a framework that extends the traditional Markov decision process (MDP) to incorporate an additional constraint. A CMDP is defined by the tuple $\langle \mathcal{S}, \mathcal{A}, \rho, P, R, C, \gamma\rangle$: state space $\mathcal{S}$, action space $\mathcal{A}$, initial state distribution $\rho$, transition probability $P$, reward function $R$, cost function $C$, and discount factor $\gamma$. The expected reward sum $J(\pi)$ can be written in the aforementioned terms as follows:
\begin{equation}
    \begin{split}
        J(\pi) & \coloneqq \mathbb{E}_{\pi} \left[\sum_{t=0}^{\infty}\gamma^t R(s_t,a_t)\right],
    \end{split}
\end{equation}
where $a_t \sim \pi(\cdot|s_t)$ and $s_{t+1} \sim P(\cdot|s_t, a_t)$. Similarly, to define constraints, the expected cost sum can be expressed as follows:
\begin{equation}
    C_\pi \coloneqq \mathbb{E}_\pi \left[\sum_{t=0}^{\infty}\gamma^t C(s_t,a_t) \right].
\end{equation}
Then the objective of safe RL can be represented as follows:
\begin{equation}
    \begin{split}
        \text{maximize}_{\pi} ~ J(\pi) \text{ s.t. } C_\pi \le \frac {d}{1-\gamma}, \label{safeobj} 
    \end{split}
\end{equation}
with the constraint threshold $d$.

\subsection{Constraint reward}
\label{corsec}
Constraint reward (CoR) is an additional objective term that assesses the relative distance of an agent state between two sets of state data \cite{MixGAIL}. By utilizing two disparate sets of states, denoted as $S_A$ and $S_B$ respectively, the agent can estimate its performance relative to these two sets of demonstrations. If the distance between the agent's state and the first set of states, $S_A$, is less than the distance to the other set of states, $S_B$, the CoR value exceeds 0.5. In contrast, when the agent's state is closer to $S_B$ than $S_A$, the CoR is reduced to below 0.5. In the prior work \cite{MixGAIL}, by defining $S_A$ as the collection of states associated with expert performance and $S_B$ as those corresponding to suboptimal or negative behavior, such as random policy, the CoR enables the training of agents to emulate expert trajectories over undesirable ones. For the state $s$, the CoR is defined as follows:
\begin{equation}
    \begin{split}
        \text{CoR}(s, S_A, S_B) =& \frac{\left(1+\frac{\Delta_{A}}{\alpha}\right)^{-\frac{\alpha+1}{2}}}{\left(1+\frac{\Delta_{A}}{\alpha}\right)^{-\frac{\alpha+1}{2}} + \left(1+\frac{\Delta_{B}}{\alpha}\right)^{-\frac{\alpha+1}{2}}}, \\ \label{cor}
        ~ \\
        \Delta_{A} =& \sqrt{\frac{1}{\lvert S_A \rvert} \sum_{s_a \in S_A} \lVert{s-s_a}\rVert_2^2}, \\
        \Delta_{B} =& \sqrt{\frac{1}{\lvert S_B \rvert} \sum_{s_b \in S_B} \lVert{s-s_b}\rVert_2^2}, \\
    \end{split}
\end{equation}
where $\lVert \cdot \rVert_2$ is the $l_2$ norm, and $\alpha$ refers to a hyperparameter used to regulate the sensitivity of CoR.

\section{Safe CoR}
The goal of this work is to combine the strengths of imitation learning (IL) with those of safe reinforcement learning (safe RL) by utilizing expert demonstrations. The most straightforward method of combining IL and RL is to redesign the actor's objective by incorporating an imitation learning term, such as log-likelihood probability, $\mathbb{E}_{(s,a) \sim D}\left[\log \pi(a \middle | s) \right]$, where $D=\{s_0, a_0, \ldots, s_N, a_N\}$ is a dataset of expert trajectories, as in \cite{bcsac}. However, challenges arise when applying this approach to safe RL. Using an expert focused solely on maximizing the reward, referred to as a reward expert, can lead the agent to violate given constraints. On the other hand, an expert trained through safe RL algorithms, represented as a safe expert, might suffer from the drawback of low reward, despite directly optimizing the constraint. In other words, relying solely on each type of expert does not align with the ideal framework we aim to build.

One approach to tackle these challenges is to utilize both demonstrations. In scenarios where safety is assured, the agent is encouraged to prioritize the influence of the reward expert over the safe expert for higher rewards. Conversely, when the agent struggles to adhere to a given constraint, it can be directed to emulate the behavior of the safe expert rather than the reward expert. Through this strategy, the agent can be steered towards an optimal balance between the guidance provided by the two experts. Building upon the foundational principles outlined in the preceding sections, constraint reward (CoR) can serve as a guidance. $\text{CoR}(s_\pi, S_{re}, S_{se})$, where $S_{re}$ and $S_{se}$ refer to the sets of reward expert's and safe expert's demonstrations, respectively, allows us to evaluate the relative distance between the two experts. As the agent's state aligns with the states from the reward expert, the CoR value increases. On the other hand, it decreases as the agent's states converge towards that of the safe expert.
Thus, the CoR can be employed as an augmented reward component with the coefficient $\lambda$ for the objective function, as below:
\begin{equation}
    \begin{split}
        & \mathbb{E}_{\pi} \left[\sum_{t=0}^{\infty}\gamma^t \{R(s_t,a_t) + \lambda \text{CoR}(s_t,S_{re},S_{se})\} \right]. \label{rewcor}
    \end{split}
\end{equation}

While the enhanced reward objective helps the agent pursue a higher reward, it's necessary to regulate excessive guidance from the reward expert to ensure the agent's constraint satisfaction simultaneously, as previously discussed. To accomplish this goal, we can once more integrate the CoR into the constraint optimization process, thereby enforcing the stricter constraint on the agent as the CoR value increases. Finally, we redefine the safe RL problem in (\ref{safeobj}) as follows:
\begin{equation}
    \begin{split}
        \mathop{\text{maximize}}_\pi ~ & J(\pi) + \lambda_r \cdot \text{CoR}_\pi \\ \label{safecor}
        \text{s.t. } & C_{\pi} + \lambda_c \cdot \text{CoR}_\pi \le \frac {d}{1-\gamma},
    \end{split}
\end{equation}
where
\begin{equation}
    \text{CoR}_\pi \coloneqq \mathbb{E}_\pi \left[ \sum_{t=0}^{\infty} \gamma^t \text{CoR}(s_t,S_{re},S_{se}) \right], \\
\end{equation}
$\lambda_r$ and $\lambda_c$ are the risk adjustment parameters of the CoR term added to the $J(\pi)$ and $C_\pi$ respectively. $\lambda_r$ determines the significance of expert guidance relative to the original reward function. As its value increases, the optimization problem shifts towards the objective of IL. $\lambda_c$ influences the degree of constraint satisfaction, leading to an augmented cost function that encourages the agent to adopt a more conservative approach. Notably, an excessively large value can fundamentally alter the original optimization problem, as it might be equivalent to reducing the constraint threshold. Therefore, to ensure stable training, we assign values of 0.1 to $\lambda_r$ and 0.01 to $\lambda_c$.

\section{Experimental Setup}

\begin{figure}[htb!]
    \vspace{0.15cm}
    \centering
    \includegraphics[width=1.0\linewidth]{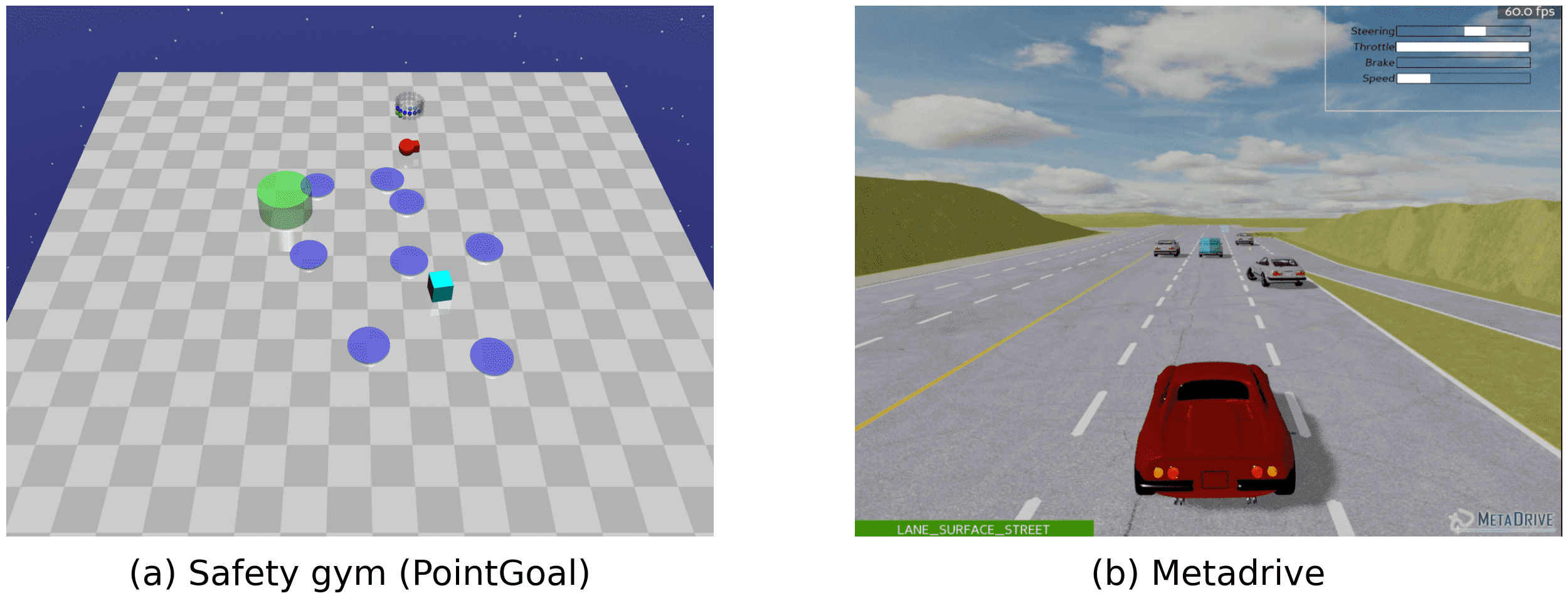}
    \includegraphics[width=1.0\linewidth]{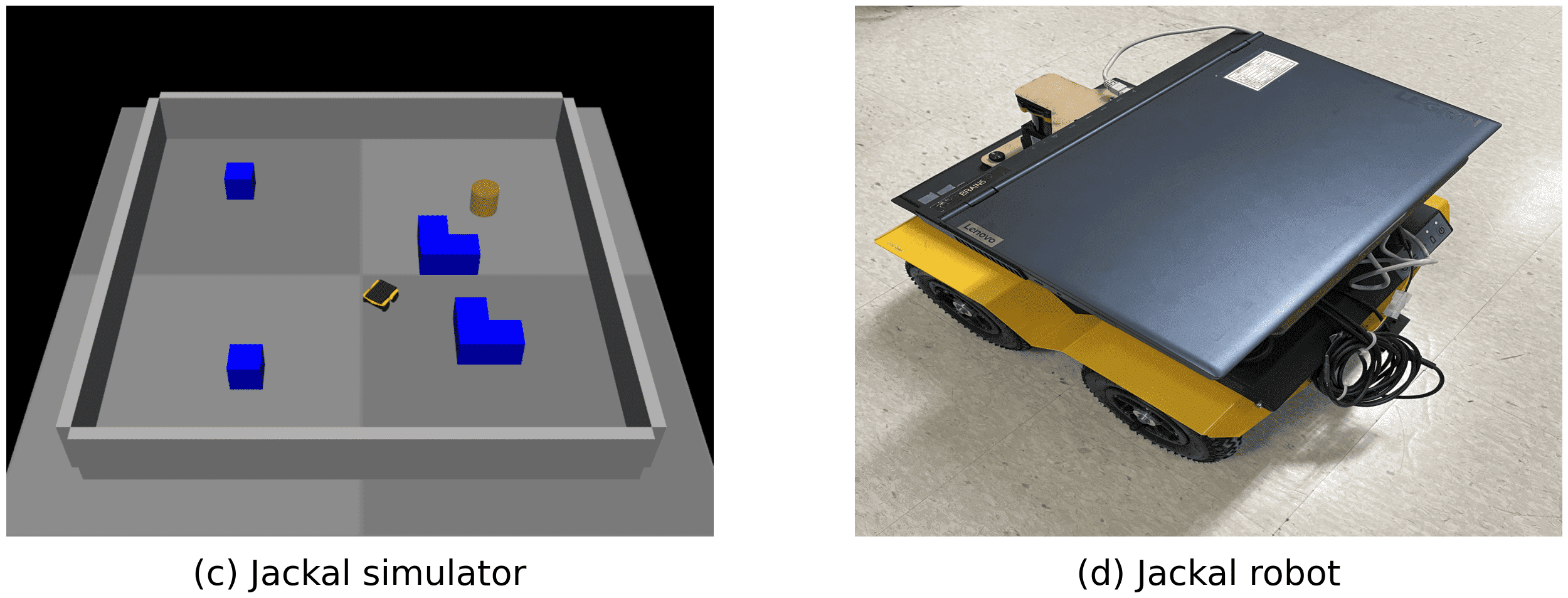}
    \caption{Experiment environments}
    \label{experiments}
\end{figure}

In this section, we detail the experimental setup by deploying our proposed framework across a combination of simulator environments and real-world platforms. Specifically, we employ two simulators, safety gym \cite{safetygym} and metadrive \cite{metadrive}, alongside the real-world Jackal platform \cite{TRC}. In the context of safety gym, we conduct comparative analyses among five safe RL algorithms, including their integration with the proposed framework. We apply PPO-Lagrangian \cite{safetygym} and WCSAC \cite{WCSAC} as representatives of Lagrangian-based methods, and utilize CPO \cite{CPO}, OffTRC \cite{OffTRC}, and SDAC \cite{SDAC} as examples of trust-region-based methods. However, due to suboptimal performance observed with Lagrangian-based methods in the Jackal simulator, the real-world Jackal platform, and the metadrive simulator, our investigation focuses solely on three trust-region-based methods. For experiments conducted in the metadrive environment, we extend our comparison to include BC-SAC \cite{bcsac}.

\begin{figure*}[t!]
    \vspace{0.15cm}
    \centering
    \includegraphics[width=0.23\linewidth]{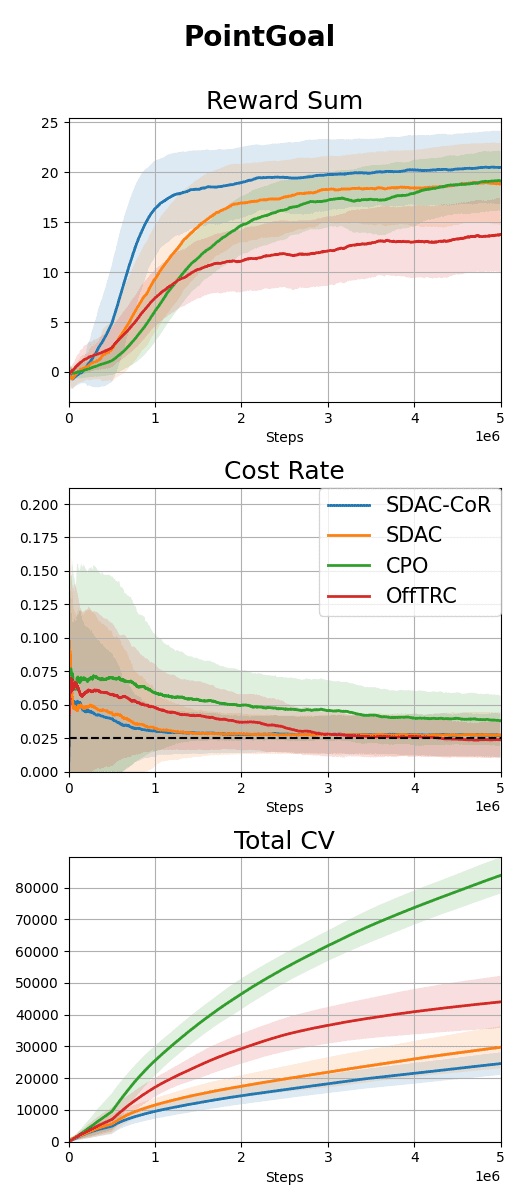}
    \includegraphics[width=0.23\linewidth]{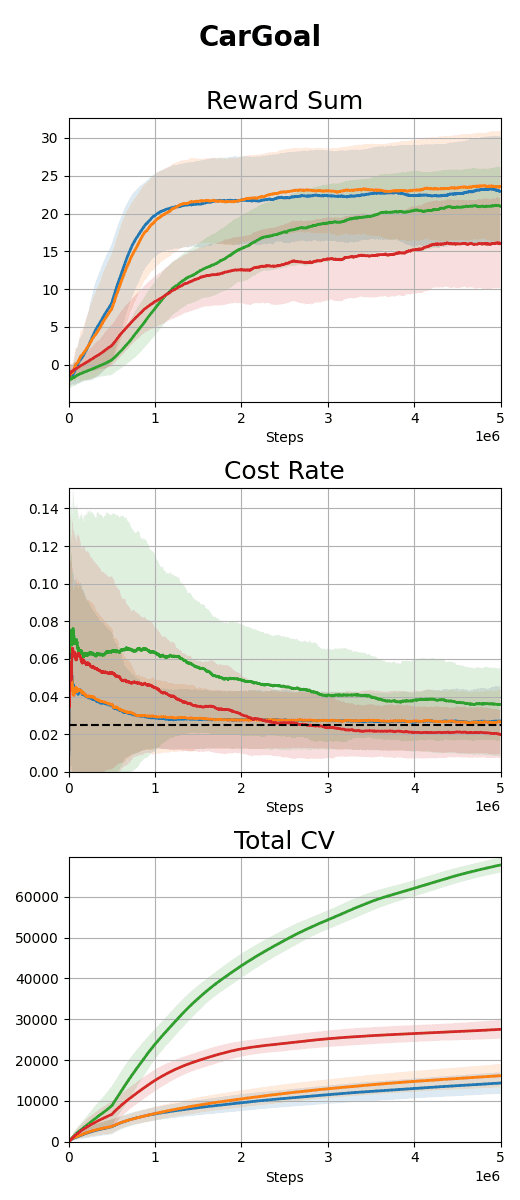}
    \includegraphics[width=0.23\linewidth]{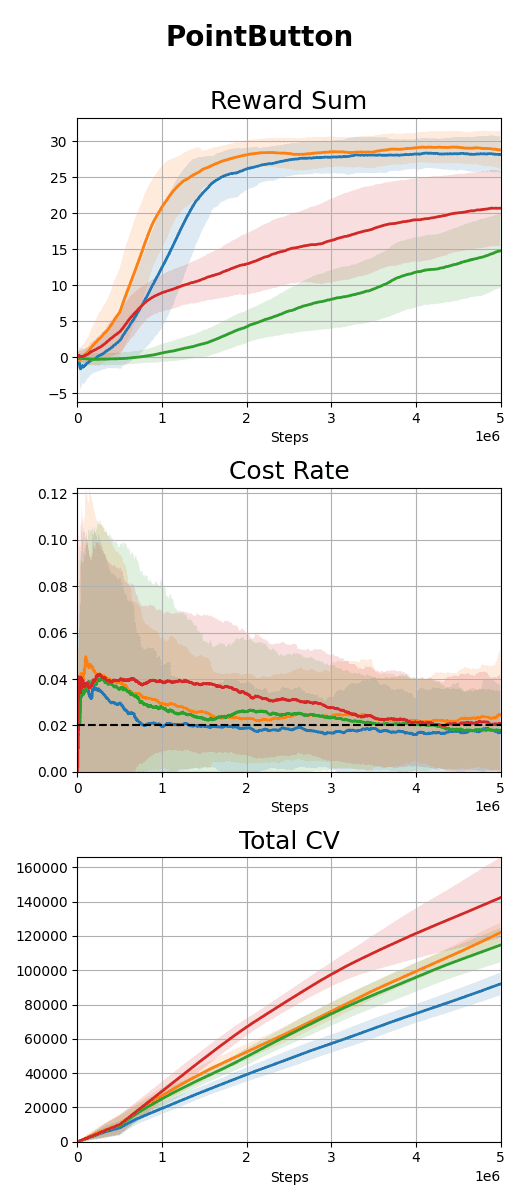}
    \includegraphics[width=0.23\linewidth]{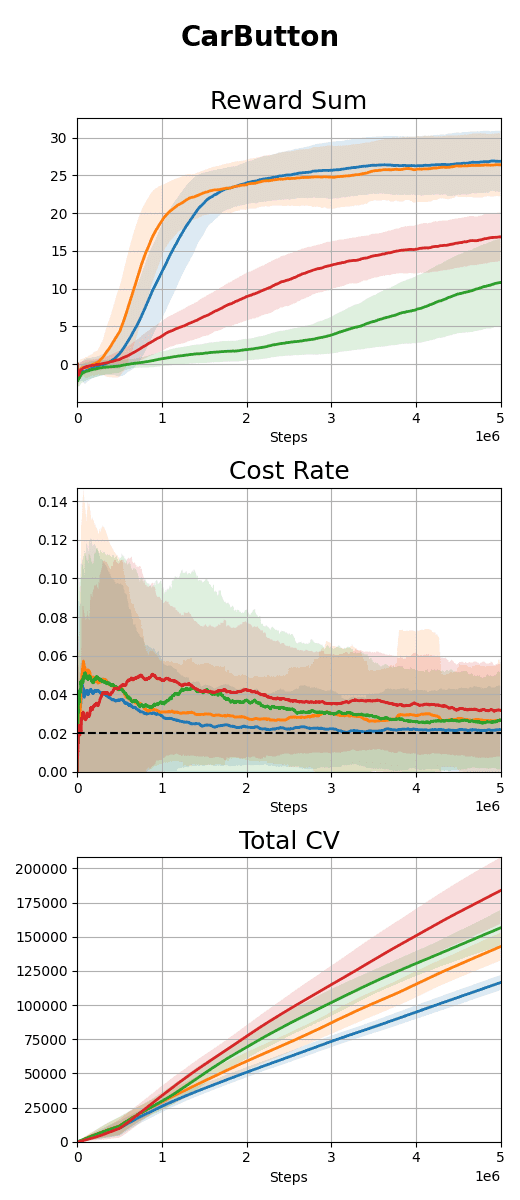}
    \caption{Safety gym results. The cost rate refers to the average cost per step and the dashed line indicates the constraint threshold.}
    \label{safetygymfig}
\end{figure*}

\subsection{Model design and hyperparameters}
In the construction of our model, we integrate a Gaussian policy network consisting of two hidden layers. In the case of the metadrive, we utilize the hidden layer dimension of 256. In the case of the safety gym and Jackal platform, environments necessitate a higher dimensional setting of 512. This architectural decision extends to the value network, which mirrors the policy network's structure and size. The networks employ the ReLU function as their activation layer.

For the model parameters, we propose to use the discount factor of 0.99, the maximum KL divergence of 0.001 per policy update step, and the learning rate of 0.0003. In the case of OffTRC, we established the risk level of CVaR as 0.25, to distinguish it from CPO. However, due to its poor performance at a risk level of 0.25, we assigned a risk level of 1.0 for SDAC. A detailed comparison between these variations will be conducted in Section \ref{ablation}. The sensitivity parameter ($\alpha$) of CoR is adjusted to 3.0.

\subsection{Safety gym}
The safety gym\cite{safetygym} is an environment that provides several tasks for testing safe RL algorithms. Unlike typical RL environments, the safety gym provides a cost function from the environment, which is the measurement of how much the agent is acting safely. The predefined robots (point and car) conduct two tasks: the goal and the button. In the goal task, an agent tries to reach a randomly spawned goal point without going through eight hazard areas. The button task requires an agent to press a designated one among four buttons while avoiding five hazard areas and four moving obstacles. For both environments, the environmental settings including the reward and cost function are the same as in \cite{SDAC}. In all tasks, the number of constraint violations (CV) is counted when the agent enters hazard zones or collides with obstacles. The constraint threshold $d$ is configured at 0.025 for the goal task and 0.02 for the button task. The description image of the environment is shown in Figure \ref{experiments}-(a).

\subsection{Metadrive}
The metadrive simulator, an autonomous driving platform, challenges an agent to navigate to a destination without deviating from the road, colliding with objects, or exceeding a maximum speed which we set to 30km/s for stable training. The simulation utilizes a state representation that includes LiDAR feedback, driving metrics, and navigation data, with actions defined by continuous variables for steering angle and acceleration. The reward function $R(s_t, a_t)$ is composed of two key components: achieving higher speed and successfully navigating through prescribed waypoints. Conversely, the cost function $C(s_t, a_t)$ is designed to increment by 1 for any instances of generating unsafe scenarios, such as off-road incidents or collisions. Additionally, we define the score function $\Theta(s_t, a_t)$ using the coefficient $l_c$($= 5$) in Eq. (\ref{score}) to assess the overall driving performance. The constraint threshold $d$ is configured at 0.02

\begin{equation}
    \Theta(s_t, a_t) \coloneqq R(s_t, a_t) - l_c 
    C(s_t, a_t). \\ \label{score}
\end{equation}

Our methodology involves training experts via TRC \cite{TRC} with different constraint thresholds. We set the constraint threshold to 0.5 for the reward expert and 0.001 for the safe expert. Evaluation during test phases measures the average sum of score, reward, violations, and success probabilities over 100 episodes, each 1,000 steps in length. Upon an off-road incident, the environment is reset, where the agent's position is reverted to the start while the step count continues from its preceding value, ensuring step continuity despite the positional reset. The simulator's layout and functionality are illustrated in Figure \ref{experiments}-(b).

\subsection{Jackal simulator and real-world platform}
For the real-world Jackal experiment, we apply the sim-to-real method to the Jackal platform, using the pretrained agents from the Jackal simulator \cite{TRC}. The Jackal simulator is an environment for training safe RL algorithms, where it utilizes the robot platform for solving the goal task of the safety gym. The state is composed of LiDAR sensor data, linear and angular velocity, as well as goal direction and distance. The overall settings of the simulator are the same as in \cite{OffTRC}. The constraint threshold $d$ is configured at 0.025. The description image of the environment is shown in Figure \ref{experiments}-(c) and (d).

\section{Results}
\subsection{Safety gym}
\begin{table}[t]
\begin{center}
    \begin{tabular}{c||cccc}
        \hline
        Algorithm & Reward & Cost($\le 25$) & CV & Total CV($\times~10^3$) \\
        \hline
        PPO-L\cite{safetygym} & 1.55 & \bfseries14.559 & \bfseries1.915 & 31.315 \\
        PPO-L+CoR & \bfseries4.203 & 21.168 & 4.23 & \bfseries22.192 \\
        \hline
        WCSAC\cite{WCSAC} & 12.744 & 38.382 & 1.08 & 32.778 \\
        WCSAC+CoR & \bfseries14.951 & \bfseries24.903 & \bfseries0.21 & \bfseries25.407 \\
        \hline
        SDAC\cite{SDAC} & 17.479 & 24.467 & 2.83 & 33.06 \\
        SDAC+CoR & \bfseries19.336 & \bfseries21.714 & \bfseries1.723 & \bfseries24.581 \\
        \hline
        CPO\cite{CPO} & \bfseries21.554 & 43.231 & 11.31 & 83.816 \\
        CPO+CoR & 19.119 & \bfseries34.951 & \bfseries5.81 & \bfseries62.776 \\
        \hline
        OffTRC\cite{OffTRC} & \bfseries16.117 & 25.253 & 2.507 & 44.016 \\
        OffTRC+CoR & 14.927 & \bfseries18.065 & \bfseries1.327 & \bfseries40.81 \\
        \hline
    \end{tabular}
\end{center}
\caption{PointGoal results. The average reward sum, cost sum, and constraint violations for each algorithm across five seeds within 1000 steps per episode are presented. Total CV refers to the sum of constraint violations in the training phase.}
\label{pointgoaltab}
\end{table}

In this subsection, our objective is to evaluate and compare various safe RL algorithms alongside their enhanced counterparts that incorporate the proposed framework, specifically within safety-gym environments. The results from each environment are illustrated in Figure \ref{safetygymfig}. Note that due to the superior performance of SDAC with the risk level of 1.0 compared to other baseline algorithms, the figure exclusively presents the results of SDAC enhanced by the safe CoR application. It suggests that the integration of the proposed framework with SDAC yields similar outcomes to the original SDAC in terms of reward sums. Nonetheless, when evaluating the cost rate and the sum of constraint violations in the training phase (total CV), the framework significantly outperforms the baseline algorithms across all environments.

Furthermore, in order to assess the versatility and effectiveness of the proposed framework, we conducted an experiment on the PointGoal task with baseline algorithms augmented by the safe CoR. As demonstrated in Table \ref{pointgoaltab}, the implementation of the proposed framework resulted in a beneficial impact on the overall performance. For both WCSAC and SDAC, the utilization of the framework led to a decrease in the sum of costs and constraint violations, while simultaneously enhancing the sum of rewards. Remarkably, while SDAC initially stood out as a state-of-the-art algorithm among the baselines, the proposed framework further enhanced its performance. For PPO-L, the safe CoR demonstrated improvements in both reward sum and total constraint violations, while still maintaining adherence to the constraint threshold. The decrease in reward sum observed in CPO with the safe CoR is considered reasonable, given that the original CPO algorithm faced challenges in satisfying constraints. However, in the case of OffTRC, the proposed framework influenced the reward sum to decrease, resulting in a considerably lower cost sum compared to the constraint threshold. Considering that OffTRC operated with a risk level of CVaR as 0.25, indicative of an already conservative training approach, the implementation of the safe CoR led the agent to adhere to excessively stringent constraint.

\begin{figure*}[htb!]
    \vspace{0.15cm}
    \begin{subfigure}{\textwidth}
        \centering
        \includegraphics[width=0.23\linewidth]{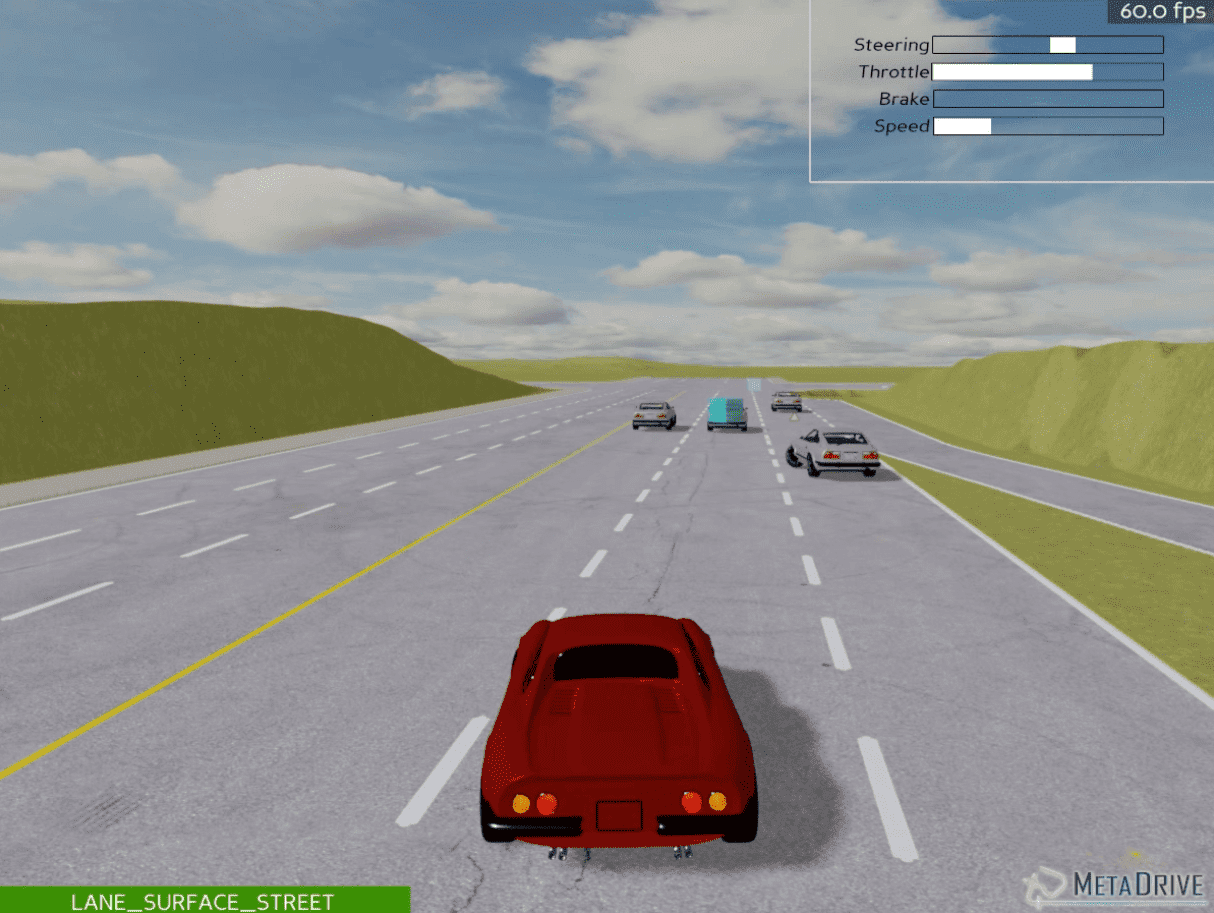}
        \includegraphics[width=0.23\linewidth]{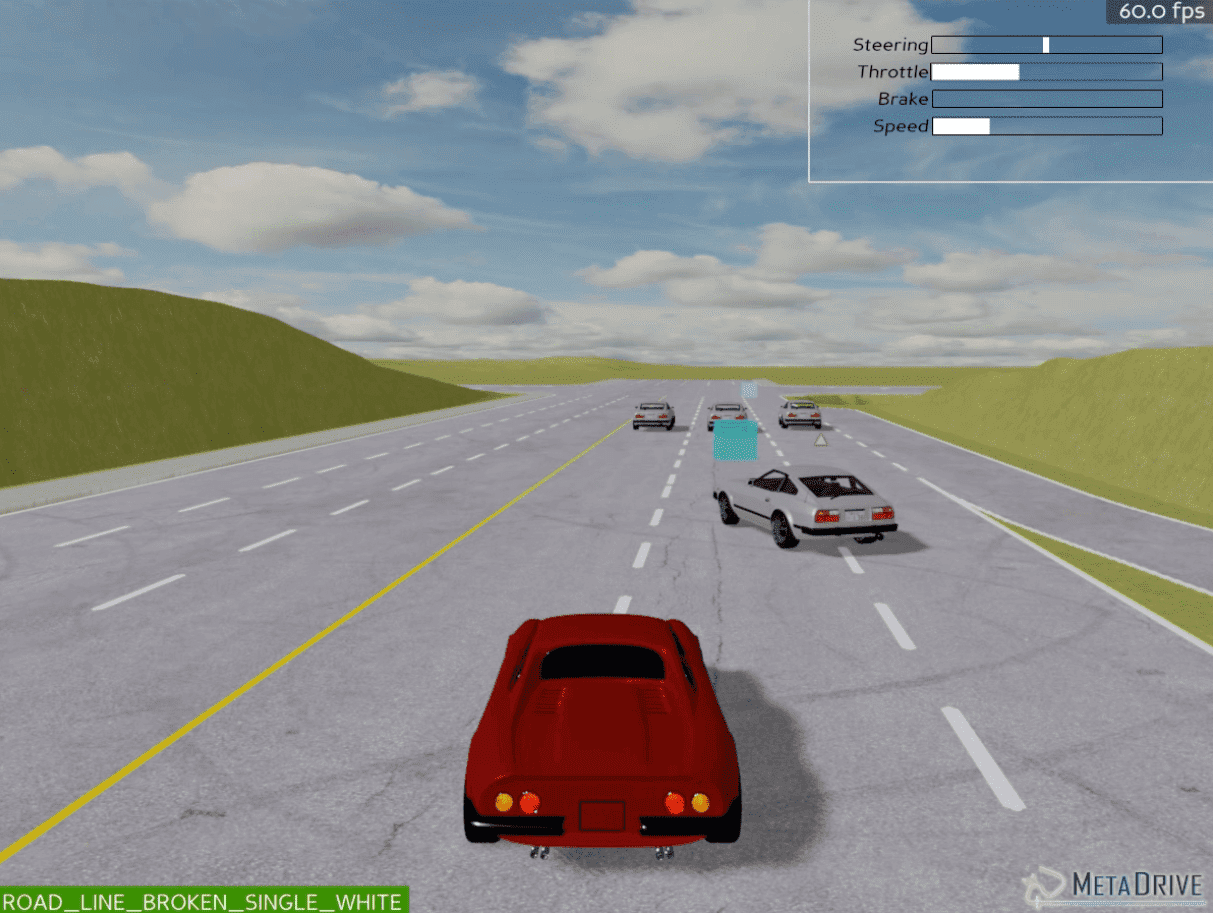}
        \includegraphics[width=0.23\linewidth]{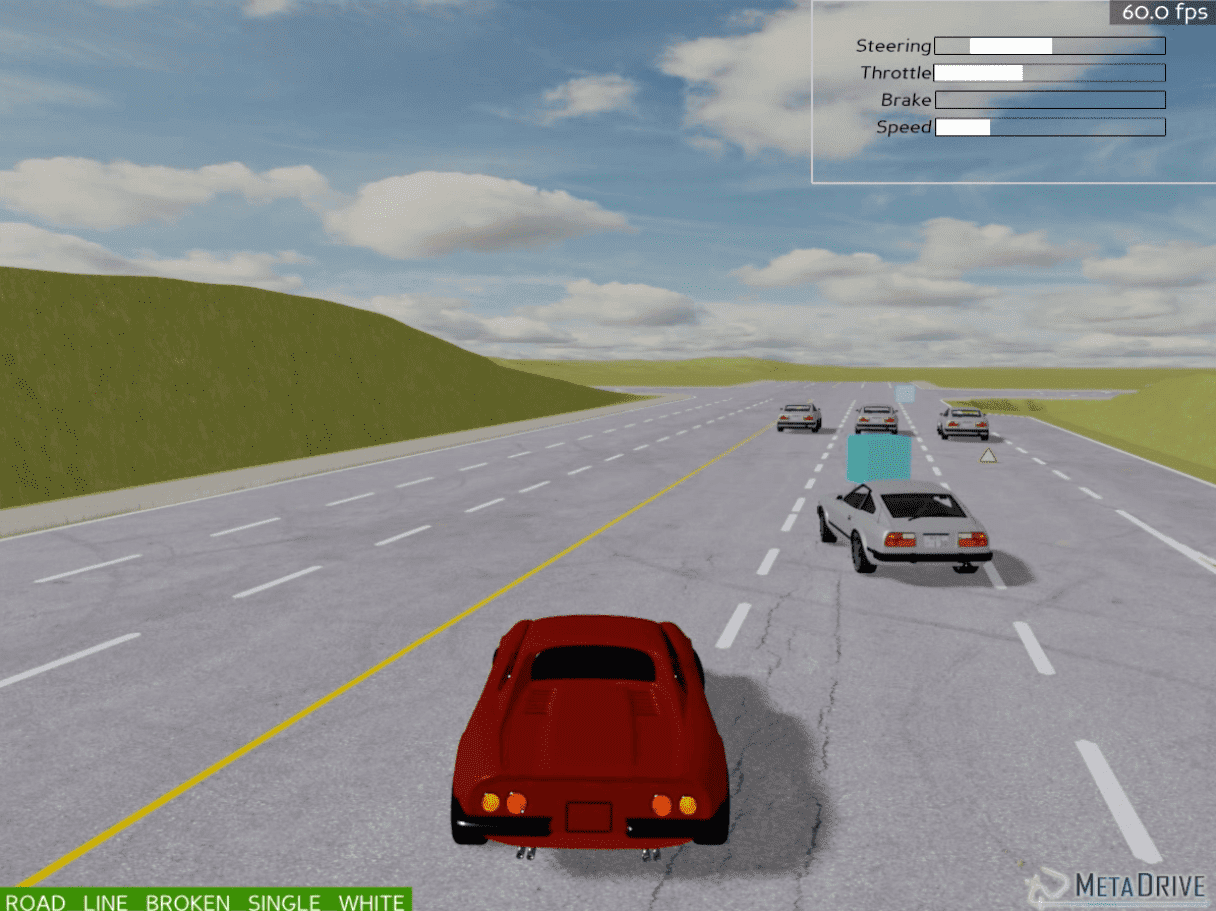}
        \includegraphics[width=0.23\linewidth]{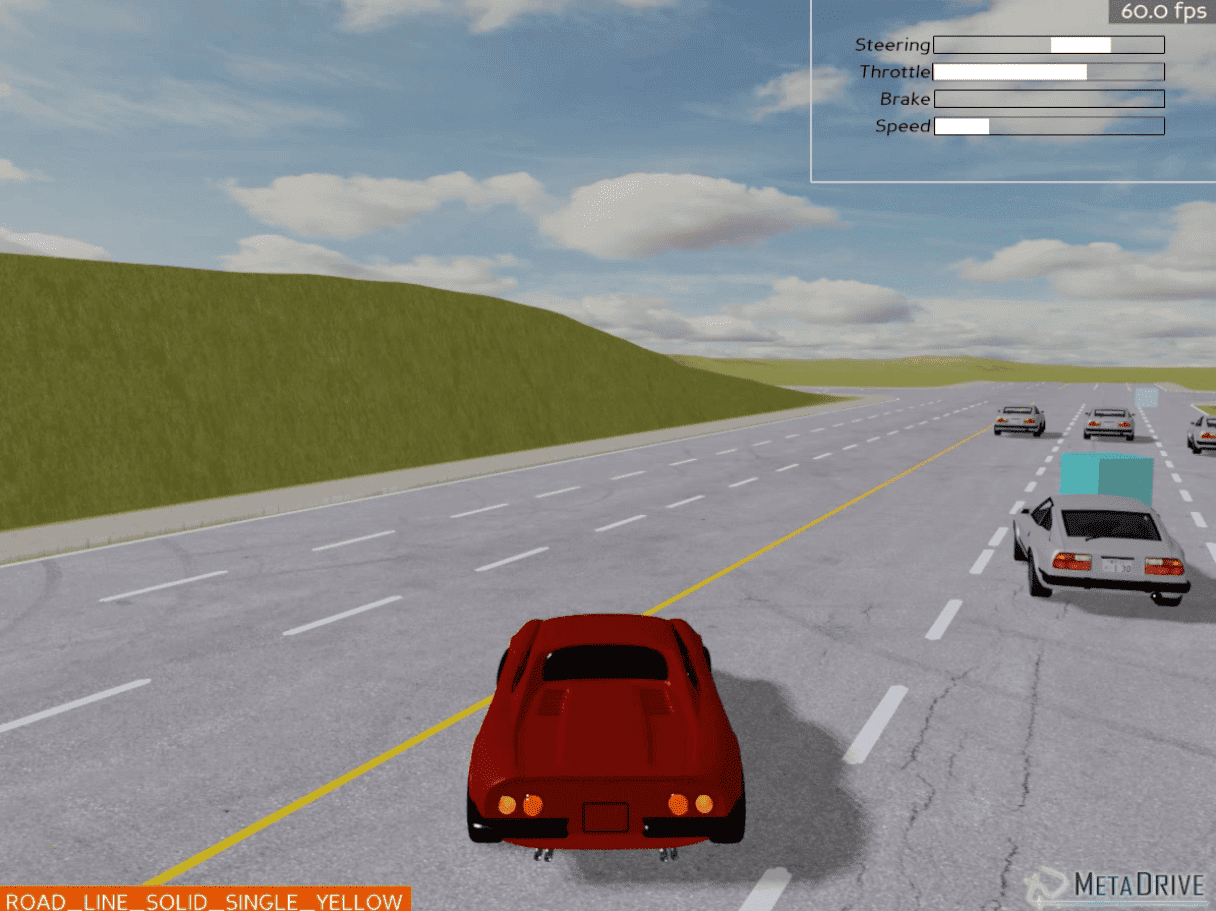}
        \caption{}
        \label{off_snapshots}
    \end{subfigure}
    
    \begin{subfigure}{\textwidth}
        \centering
        \includegraphics[width=0.23\linewidth]{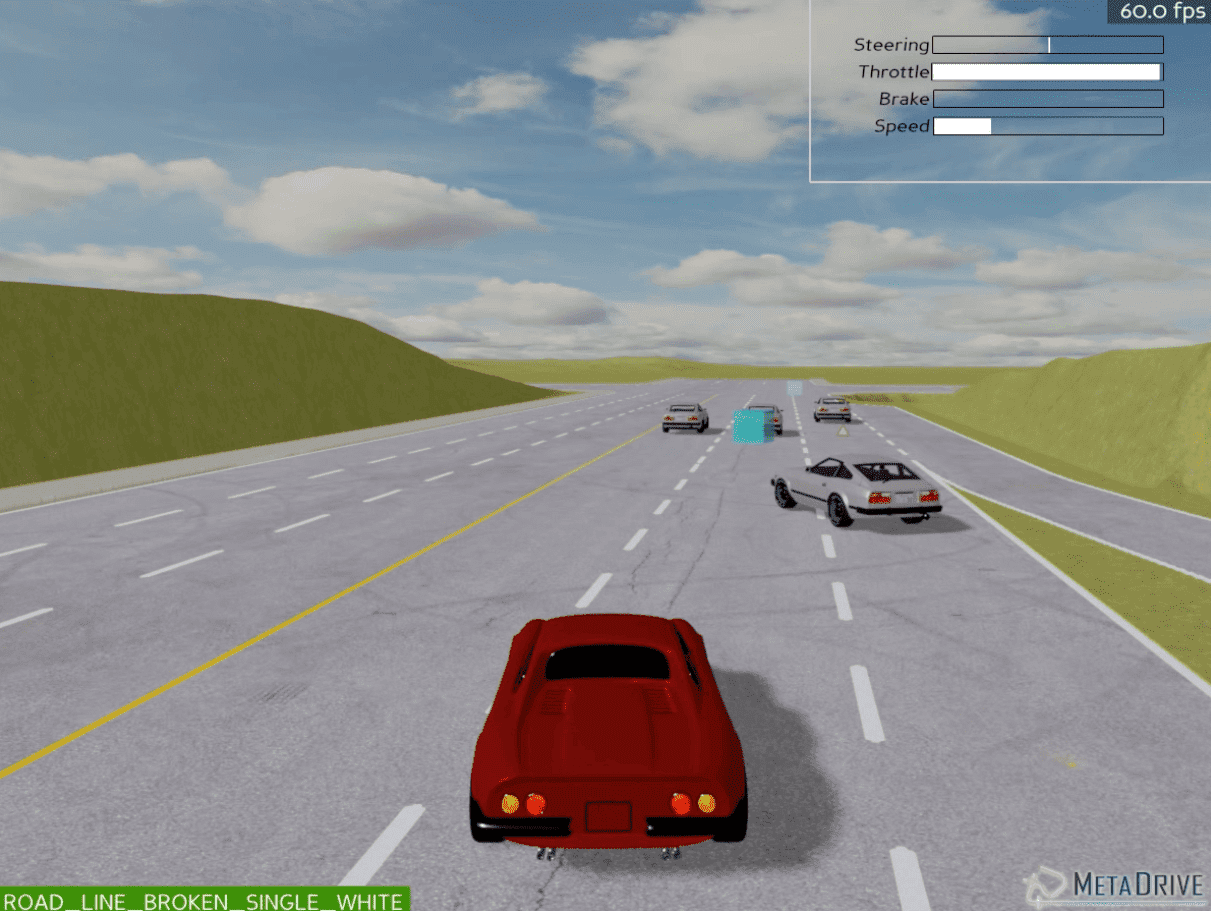}
        \includegraphics[width=0.23\linewidth]{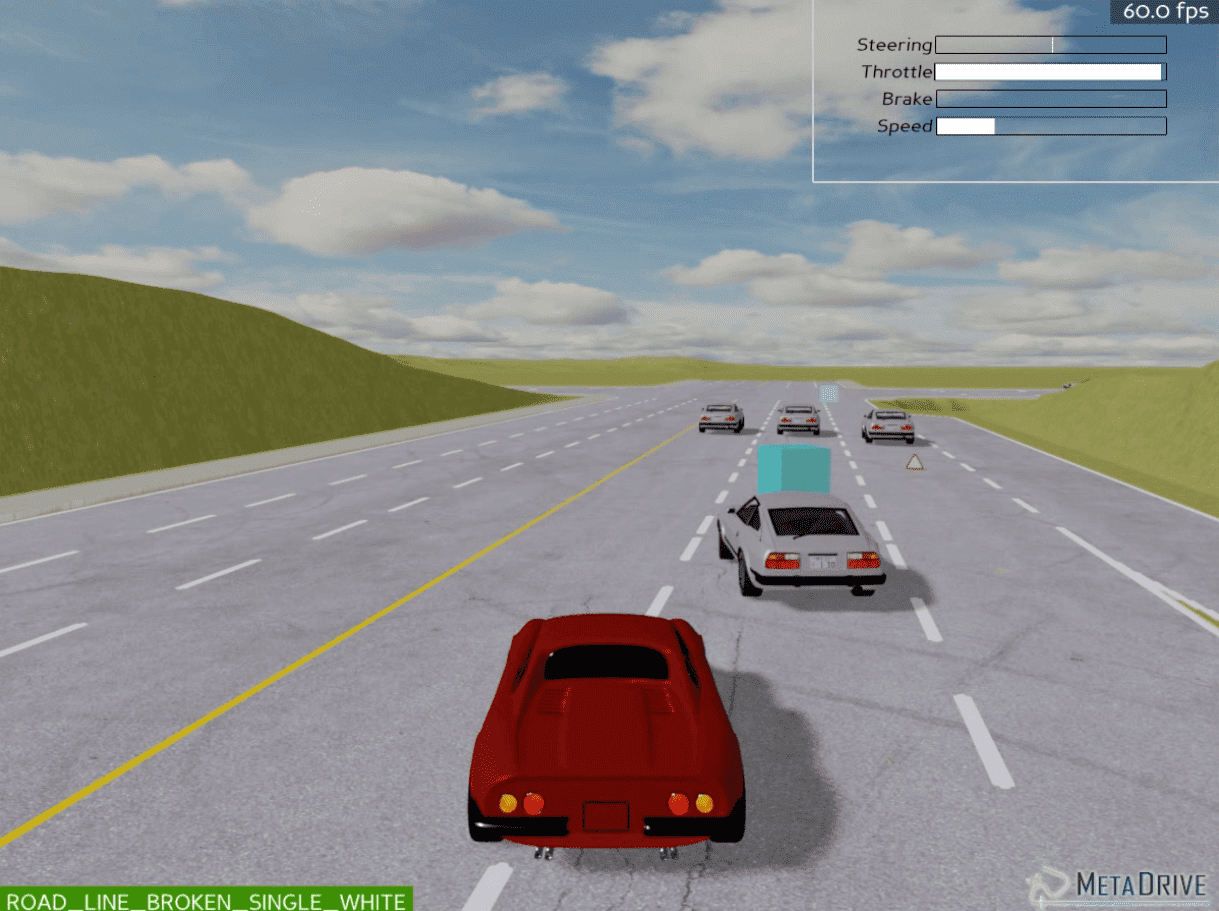}
        \includegraphics[width=0.23\linewidth]{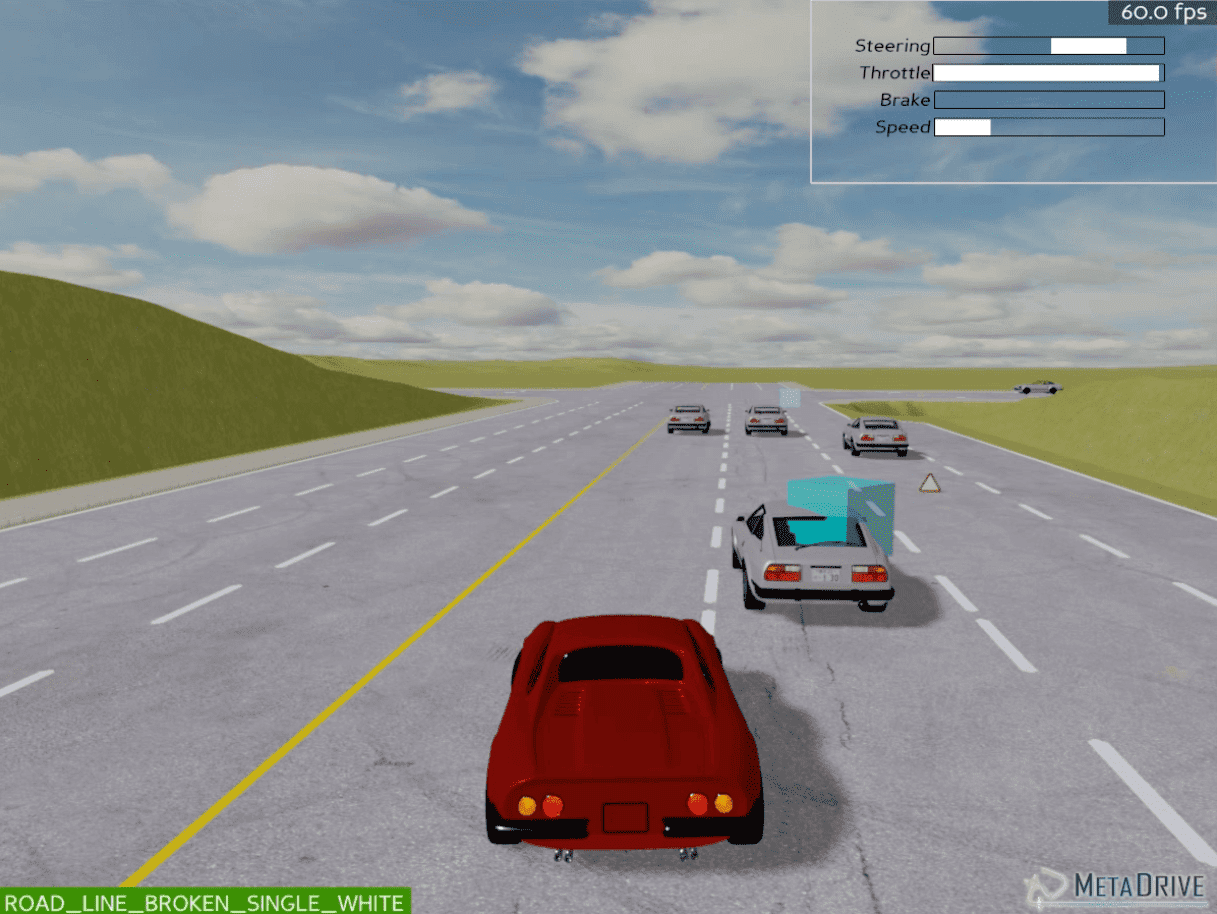}
        \includegraphics[width=0.23\linewidth]{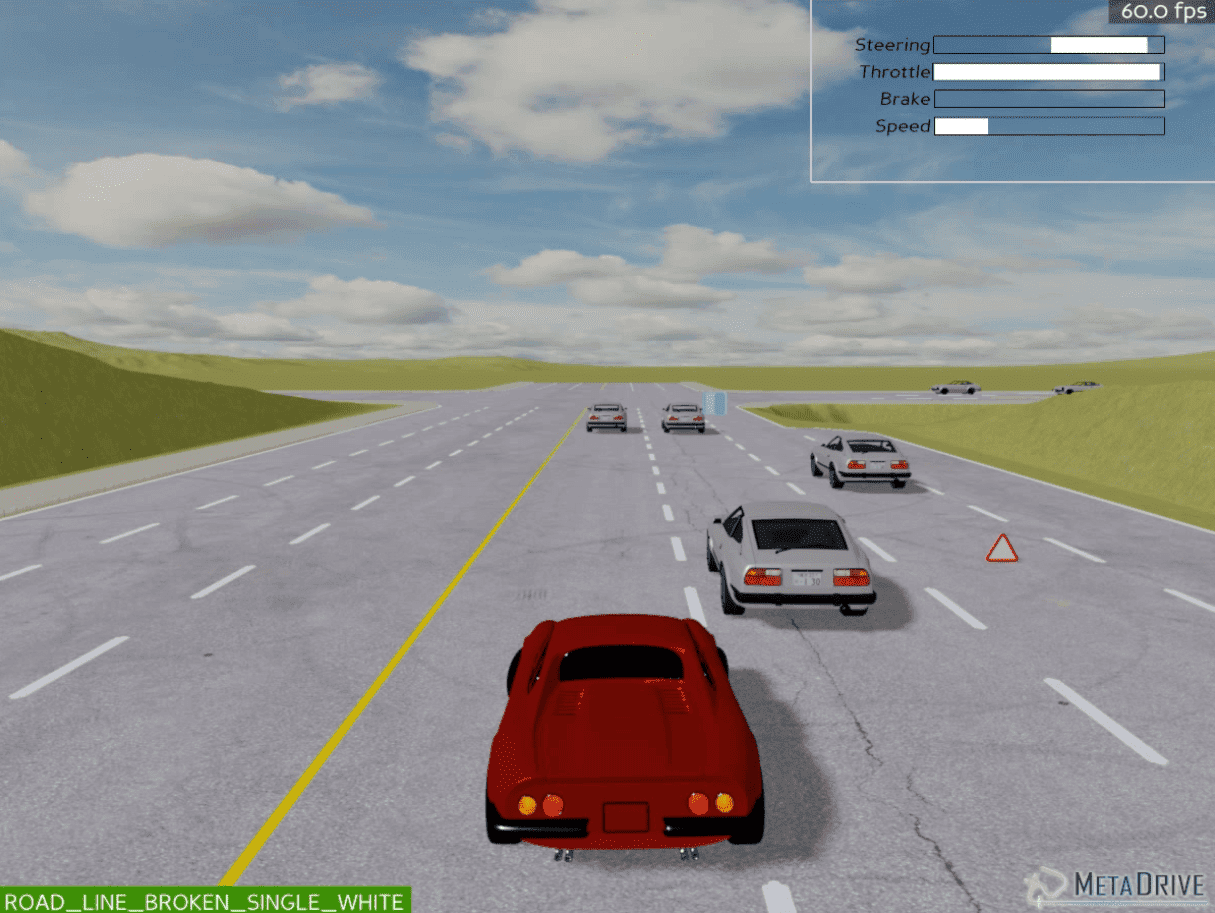}
        \caption{}
        \label{cor_snapshots}
    \end{subfigure}
    \caption{Snapshots of OffTRC (a) and OffTRC with safe CoR (b) after training in metadrive simulator}
\end{figure*}

\begin{table*}[t]
\begin{center}
        \begin{tabular}{c|c||cccccc}
        \hline
        &Algorithm & Score & Reward & Crash & Out of road & Violations & Success \\
        \hline
        \multirow{3}{*}{Expert}
        &TRC-0.5(Reward)\cite{TRC} & 564.303 & 861.530 & 25.45 & 0.14 & 25.59 & 0.97\\
        &TRC-0.001(Safe) & 780.252 &847.405 & 7.61 & 1.53 & 9.14 & 0.62\\
        &Human & 923.557 & 923.557 & 0.0 & 0.0 & 0.0 & 1.0\\
        \hline
        \multirow{2}{*}{RL}
        &SAC\cite{sac} & 76.887 &113.925 & 0.0 & 7.12 & 7.12 & 0.0\\
        &BC-SAC\cite{bcsac} & 302.547 & 334.554 & 0.78 & 5.12 & 5.56 & 0.0\\
        \hline
        \multirow{6}{*}{Safe RL}
        &SDAC & \bfseries638.554 & \bfseries763.849 & 17.1 & \bfseries5.26 & 22.36 & \bfseries0.17\\
        &SDAC+CoR & 632.830 & 753.038 & \bfseries15.2 & 6.72 & \bfseries21.92 & 0.02\\
        \cline{2-8}
        &CPO & 683.095 & 854.472 & 15.91 & 0.12 & 16.03 & 0.94\\
        &CPO+CoR & \bfseries711.330 & \bfseries855.201 & \bfseries13.6 & \bfseries0.11 & \bfseries13.71 & \bfseries0.98\\
        \cline{2-8}
        &OffTRC & 613.046 & 849.795 & 23.13 & \bfseries0.19 & 23.32 & \bfseries0.94\\
        &OffTRC+CoR & \bfseries794.472 & \bfseries882.178 & \bfseries8.64 & 0.24 & \bfseries8.88 & 0.93\\
        \hline
    \end{tabular}
\end{center}
\caption{Metadrive results. The average sum of score, reward, and violations of each algorithm and experts are presented. For each algorithm, the best performance for each element is denoted by a bold value. For TRC, we denote the value of the constraint threshold alongside its name.}
\label{metadrive_result}
\end{table*}

\subsection{Metadrive}
The final result of the simulation experiment is shown in Table \ref{metadrive_result}. The most crucial metric is the score as it indicates the agent's effort to progress along the waypoints. Additionally, the success ratio provides an overall measurement of the agent's performance in achieving the objective. The violation comprises the sum of crash and out-of-road situations. 

In the case of SAC, it exhibited minimal learning in the environment, with no instance of reaching the destination. Incorporating expert demonstrations into SAC resulted in an improved driving score. However, the agent still encountered difficulties in reaching the final goal. For safe RL algorithms, CPO exhibited superior performance. However, when applying the safe CoR, OffTRC outperforms all algorithms. The implementation of OffTRC with the framework yielded a remarkable 29.6\% enhancement in the driving score with a 61.9\% reduction in violations. Notably, the number of crashes decreased significantly. In the case of CPO, there was a 4\% improvement in the score along with a 14.5\% decrease in violations. For SDAC, the application of the framework led to lower performance. However, considering the success ratios of both versions, it appears that SDAC faced challenges during training within the simulator.

In general, safe RL agents demonstrated competent performance, and when coupled with safe CoR, they exhibited improved performance compared to the original algorithms.

\subsection{Jackal platform}
In this subsection, we conduct a comparative analysis of various safe RL algorithms and the effectiveness of the safe CoR framework applied to these algorithms, utilizing both the Jackal simulator and the real-world Jackal platform for our experiments. The outcomes from the Jackal simulator are depicted in Table \ref{jackal_simul}, and the results from the real-world Jackal platform are detailed in Table \ref{jackal_real}.

The data illustrates that the integration of the safe CoR framework with safe RL algorithms enhances performance, in terms of reducing constraint violations. Moreover, this integration leads to a reduction in the overall cost sum across most scenarios, indicating an improvement in agent safety. An exception is observed with the OffTRC algorithm in the real-world experiment, where a marginal increase in the cost sum is noted. This anomaly can be attributed to the inherent safety levels of the OffTRC method, which are already high, leaving minimal scope for further enhancements through the safe CoR framework.

Regarding the overall performances, the outcomes from the real-world Jackal platform indicate that the implementation of the proposed framework yields an improvement over conventional safe RL methods. Conversely, within the Jackal simulator environment, the deployment of CPO and OffTRC algorithms exhibits a marginal decline in score values. This phenomenon occurs as the agent opts to sacrifice potential score increases in favor of significantly reducing the cost sum.

\begin{figure*}[htb!]
    \vspace{0.15cm}
    \begin{subfigure}{\textwidth}
        \centering
        \includegraphics[width=0.24\linewidth]{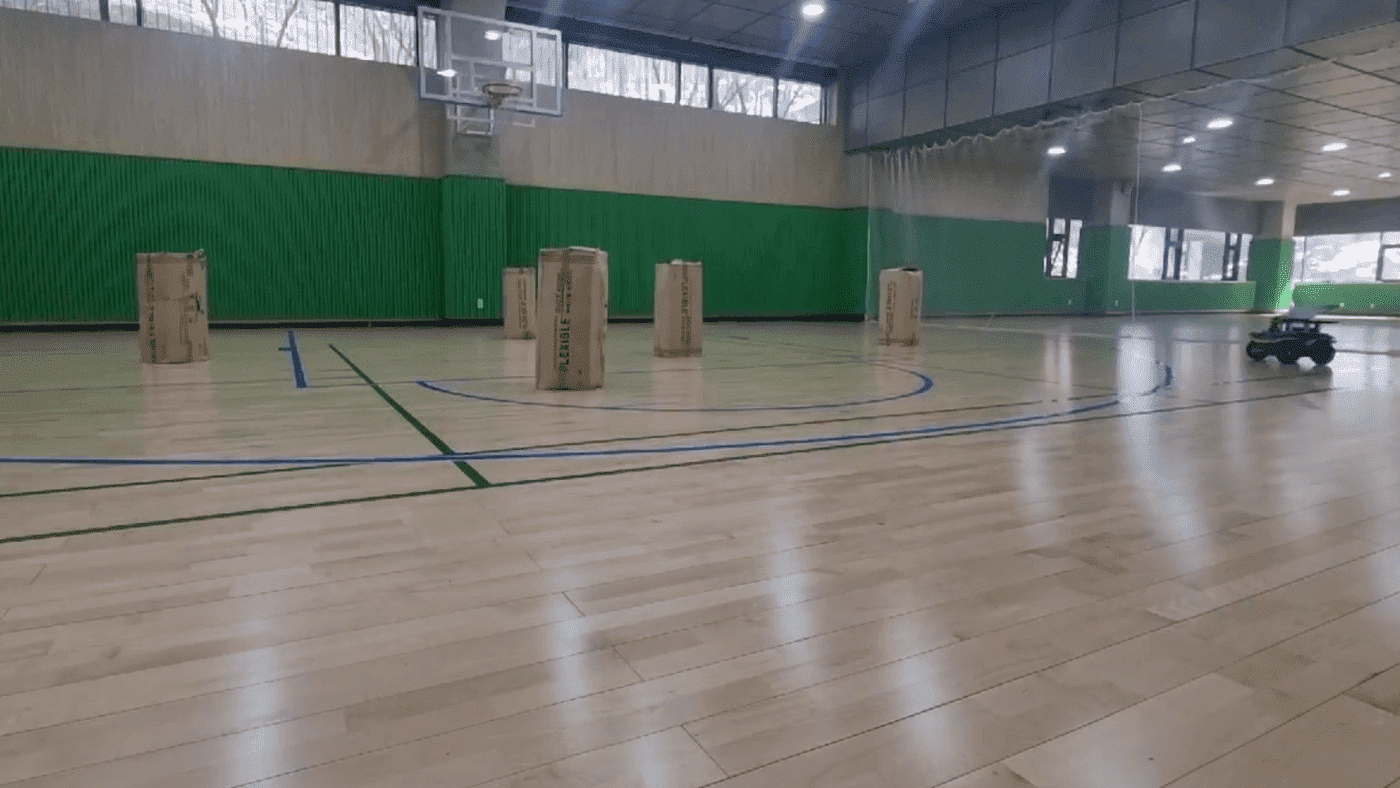}
        \includegraphics[width=0.24\linewidth]{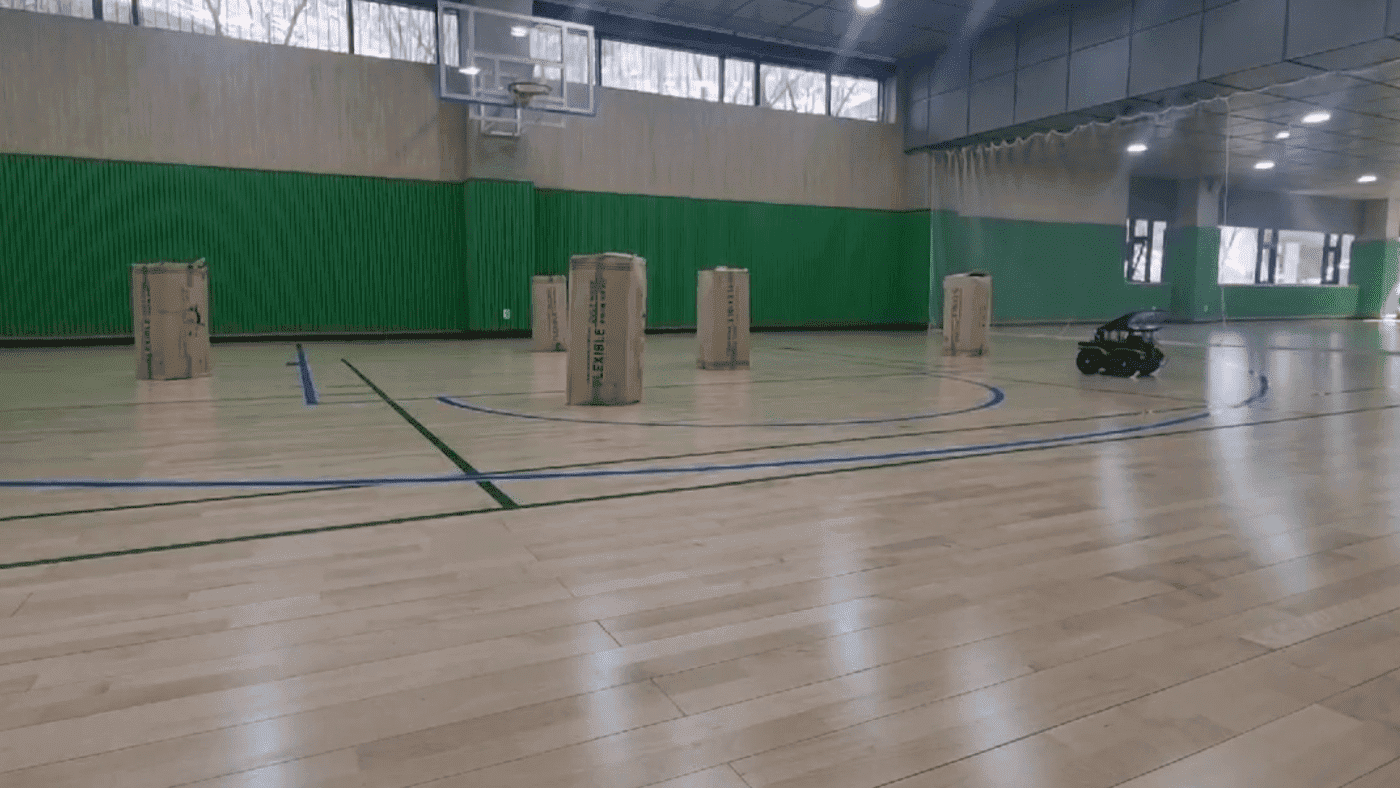}
        \includegraphics[width=0.24\linewidth]{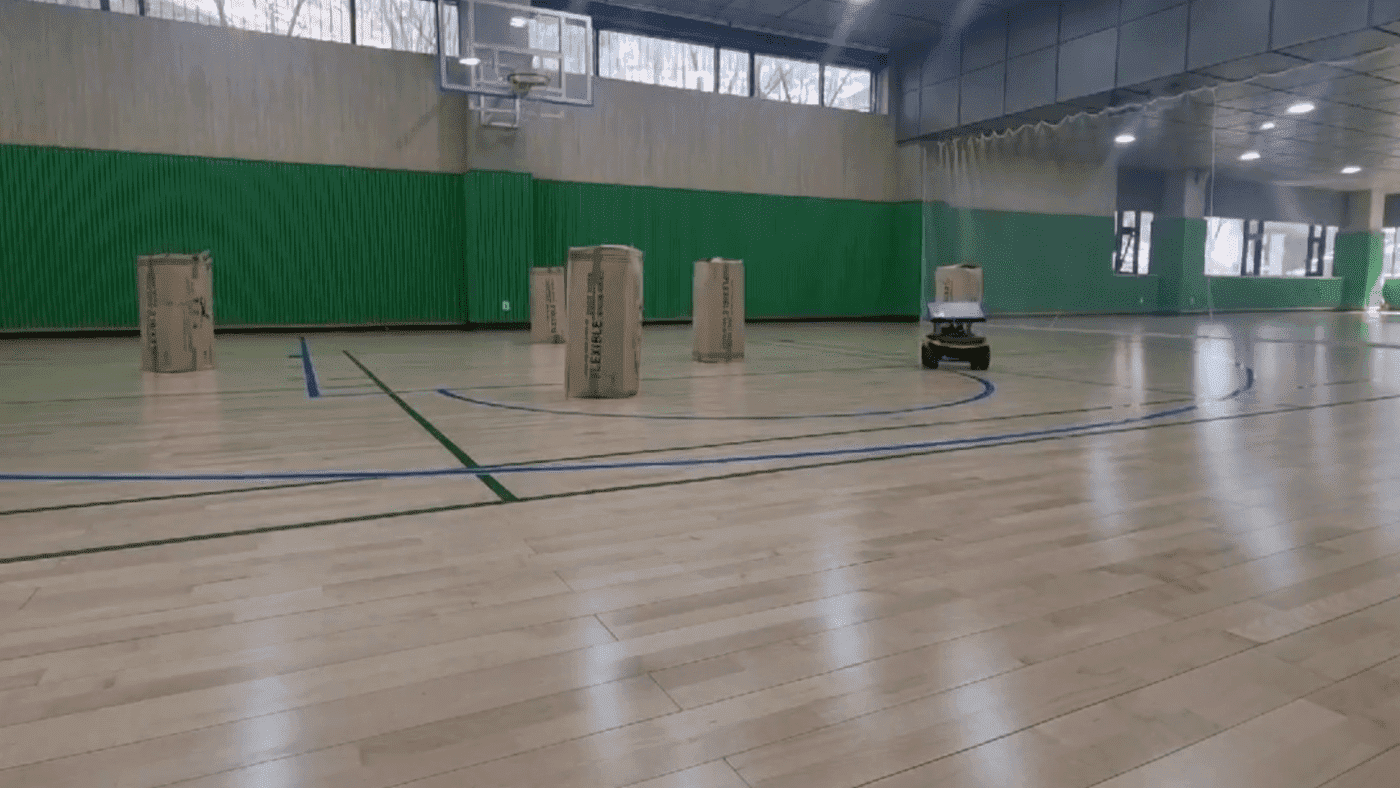}
        \includegraphics[width=0.24\linewidth]{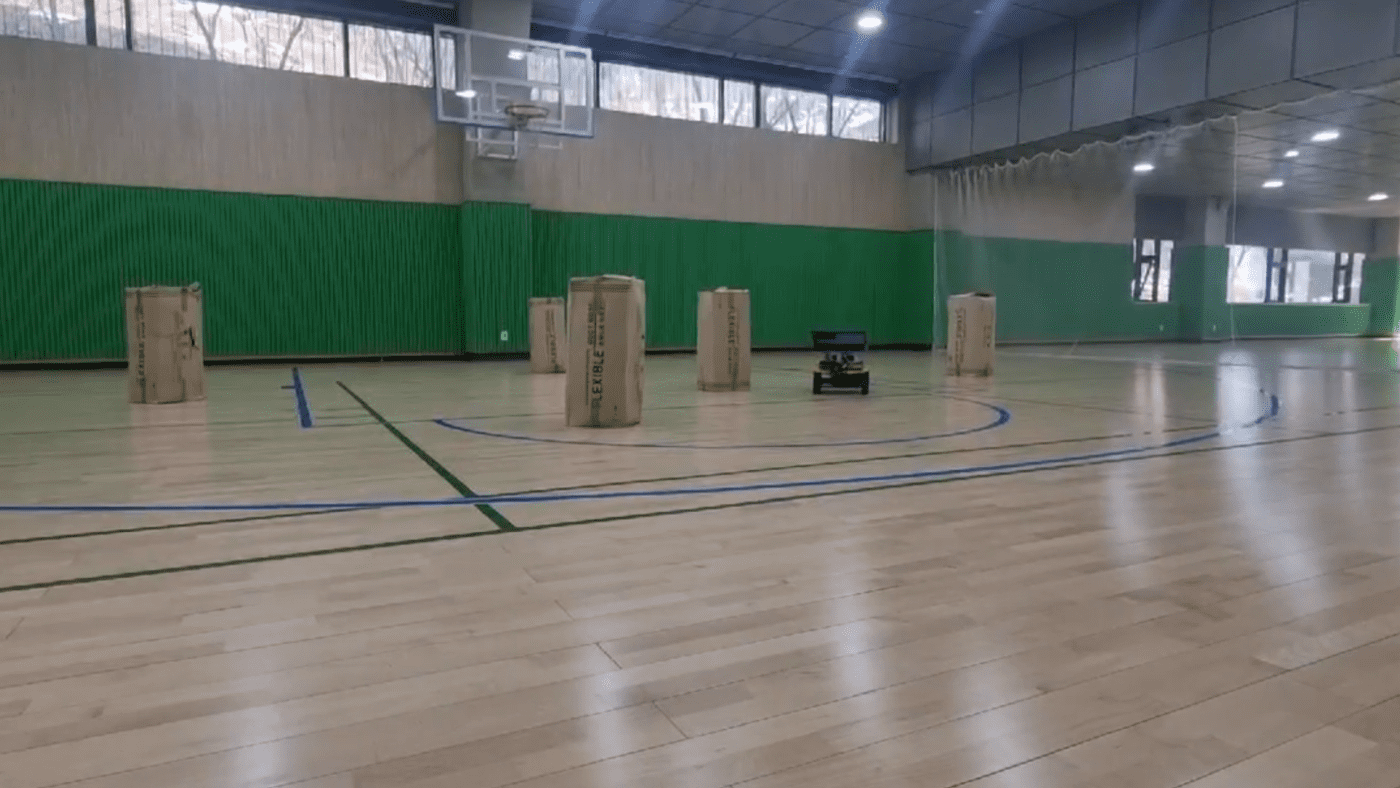}
        \caption{}
        \label{cpo_jackal}
    \end{subfigure}
    
    \begin{subfigure}{\textwidth}
        \centering
        \includegraphics[width=0.24\linewidth]{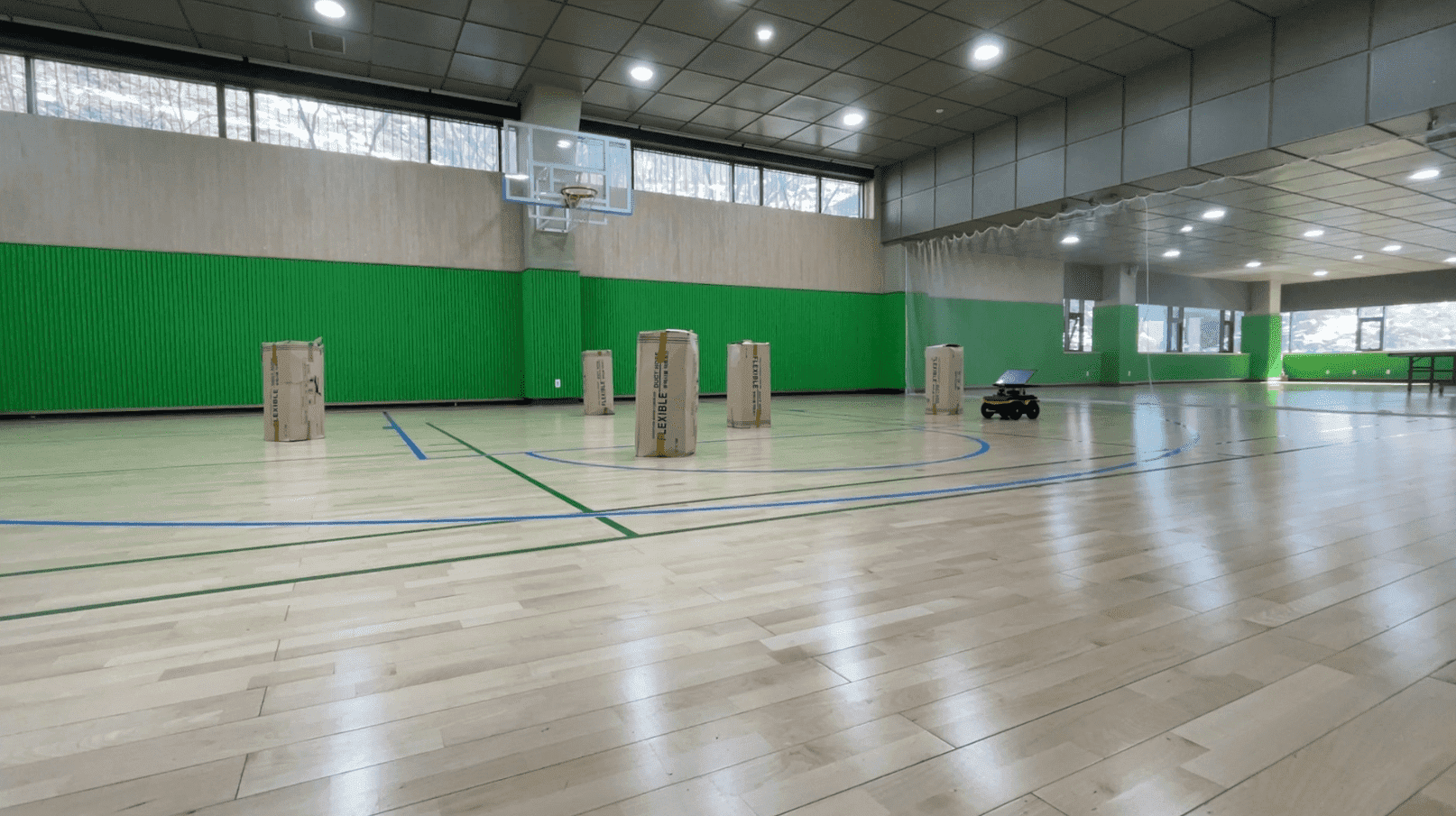}
        \includegraphics[width=0.24\linewidth]{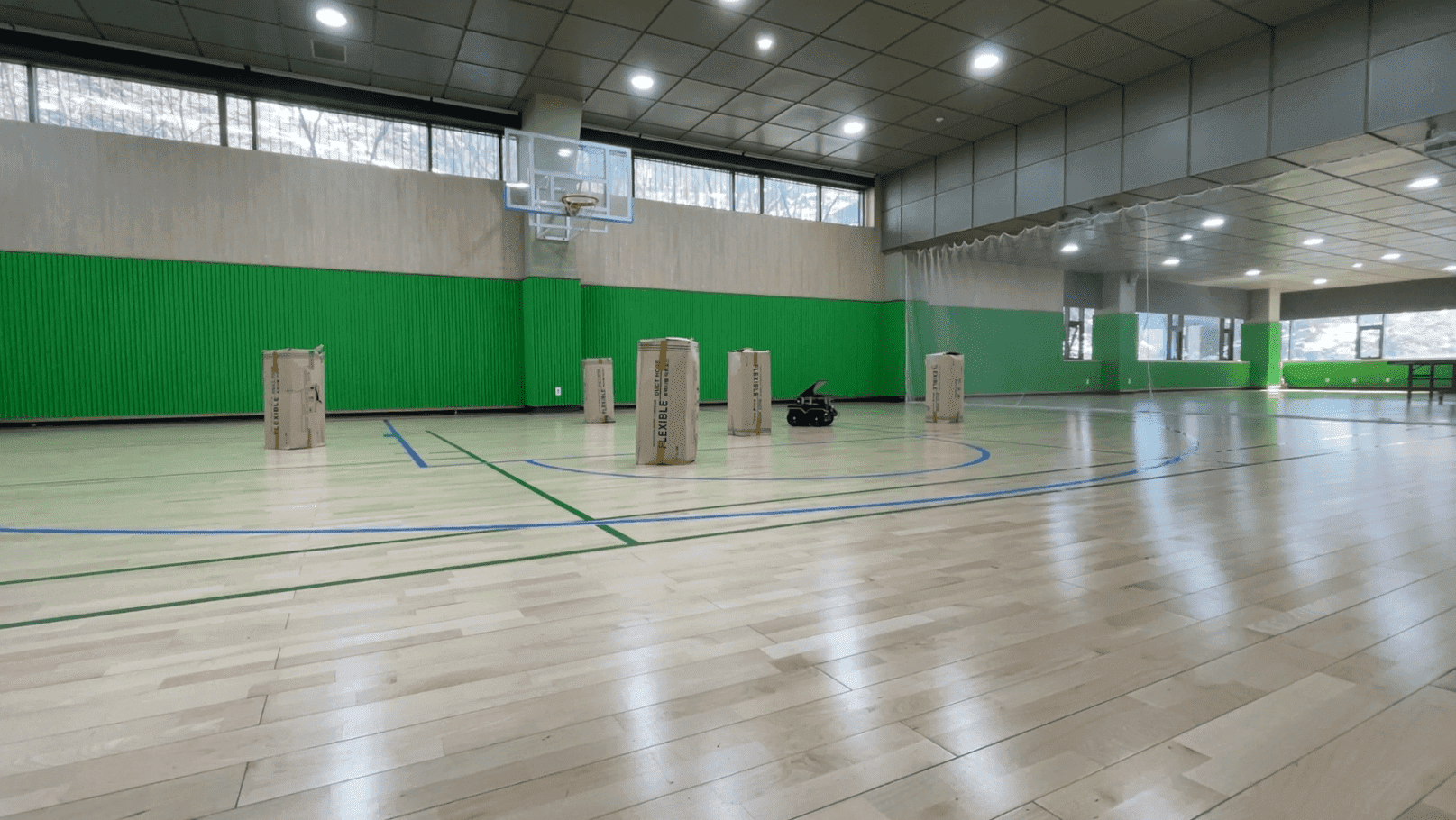}
        \includegraphics[width=0.24\linewidth]{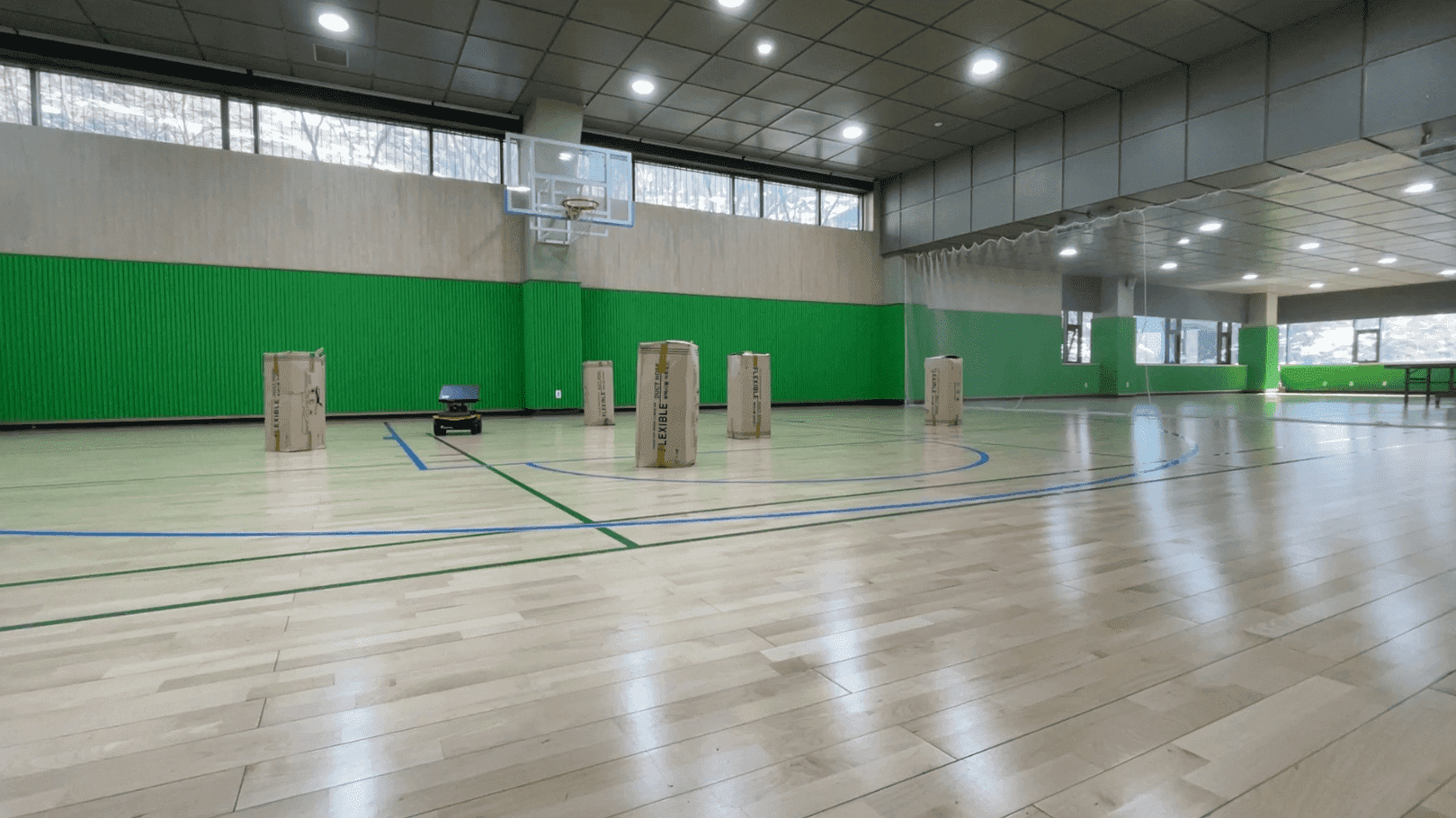}
        \includegraphics[width=0.24\linewidth]{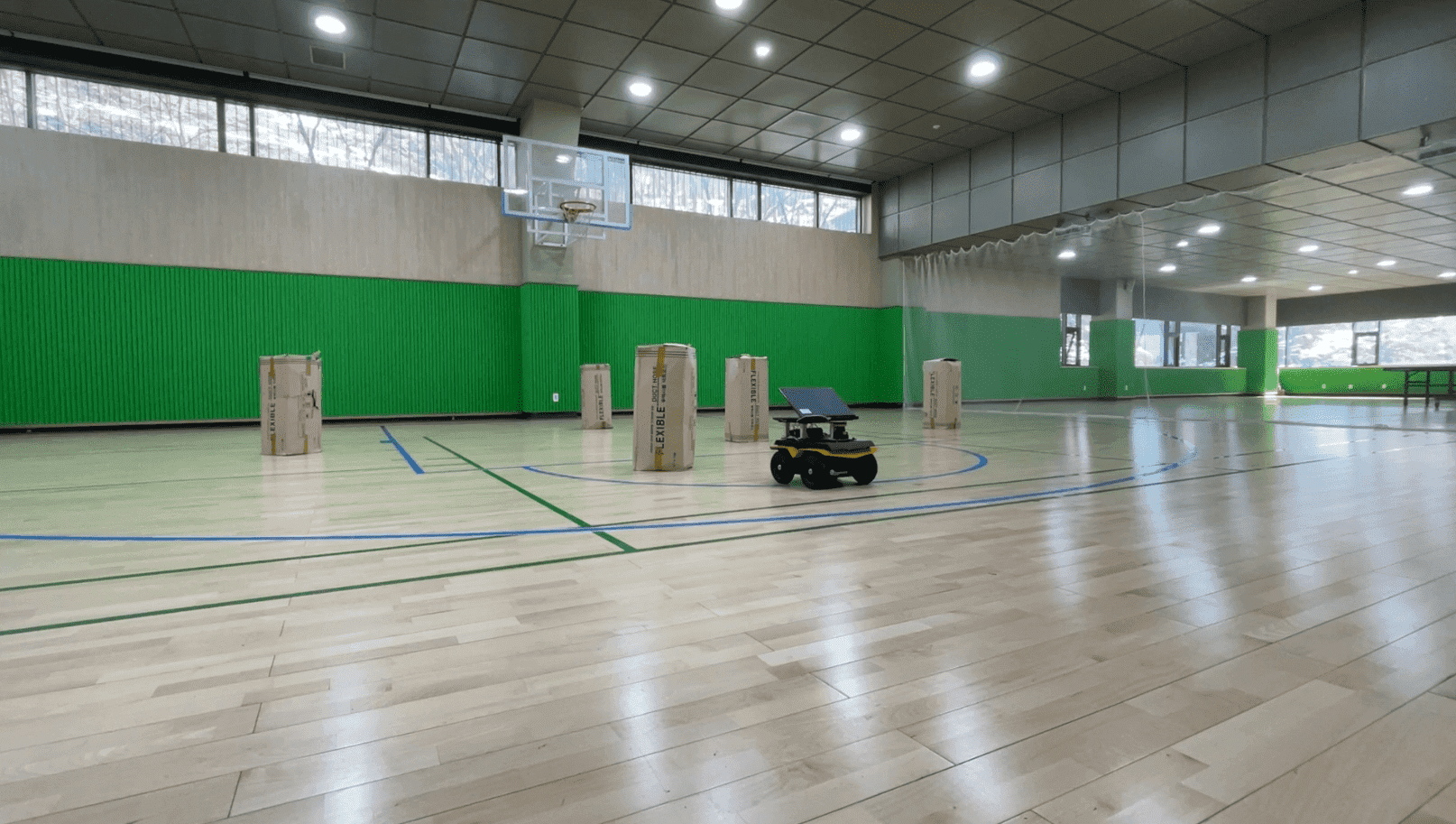}
        \caption{}
        \label{cpo_cor_jackal}
    \end{subfigure}
    \caption{Snapshots of CPO (a) and CPO with safe CoR (b) in the real-world Jackal platform.}
\end{figure*}

\begin{table}[t]
\begin{center}
    \begin{tabular}{c||ccc}
        \hline
        Algorithm & Reward & Cost($\le 25$) & CV \\
        \hline
        PPO-L & 1.464 & 24.059 & 10.825 \\
        PPO-L+CoR & \bfseries2.126 & \bfseries11.513 & \bfseries5.065 \\
        \hline
        WCSAC & 2.349 & \bfseries9.049 & \bfseries0.0 \\
        WCSAC+CoR & \bfseries6.777 & 20.313 & 14.58 \\
        \hline
        SDAC & 4.894 & 26.317 & 6.553 \\
        SDAC+CoR & \bfseries8.256 & \bfseries23.18 & \bfseries2.387 \\
        \hline
        CPO & \bfseries7.764 & 17.788 & 1.973 \\
        CPO+CoR & 7.412 & \bfseries12.94 & \bfseries1.447 \\
        \hline
        OffTRC & \bfseries12.512 & 18.039 & 1.087 \\
        OffTRC+CoR & 11.288 & \bfseries14.473 & \bfseries0.287 \\
        \hline
    \end{tabular}
\end{center}
\caption{Jackal simulator results across three seeds within 1000 steps per episode.}
\label{jackal_simul}
\end{table}

\begin{table}[t]
\begin{center}
    \begin{tabular}{c||ccc}
        \hline
        Algorithm & Reward & Cost($\le 25$) & CV \\
        \hline
        SDAC & 9.39 & 112.503 & 348.667 \\
        SDAC+CoR & \bfseries13.027 & \bfseries58.07 & \bfseries40.667 \\
        \hline
        CPO & 8.877 & 36.21 & 19.333 \\
        CPO+CoR & \bfseries16.273 & \bfseries12.6 & \bfseries0.0 \\
        \hline
        OffTRC & 21.557 & \bfseries16.48 & \bfseries0.0 \\
        OffTRC+CoR & \bfseries21.977 & 17.347 & \bfseries0.0 \\
        \hline
    \end{tabular}
\end{center}
\caption{Real-world Jackal platform results across three seeds within 1000 steps per episode.}
\label{jackal_real}
\end{table}

In summary, the results demonstrate that the application of the framework substantially reduces constraint violations and cost sums, while concurrently enhancing the score value. Consequently, it can be deduced that the proposed framework significantly augments the robustness of safe RL algorithms against environmental variations as the framework application results are superior in the real world.

\section{Ablation Study} \label{ablation}
In our ablation study, we aim to quantitatively assess the differential impacts of the CoR term on the performance and safety metrics of the navigation task using SDAC \cite{SDAC}. For appropriate comparison, we included the results obtained when setting the risk level to 0.25. Furthermore, we assessed the influence of the proposed framework in comparison to the impact of log-likelihood probability, as in \cite{bcsac}. We employed TRPO \cite{TRPO} for the reward expert and TRC \cite{TRC} for the safe expert.

\begin{table}[t]
\begin{center}
    \begin{tabular}{c||cccc}
        \hline
        Algorithm & Reward & Cost($\le 25$) & CV & Total CV($\times~10^3$) \\
        \hline
        SDAC-0.25 & 13.832 & 13.551 & 0.034 & 7.768 \\
        SDAC-1.0 & 17.479 & 24.467 & 2.83 & 33.06 \\
        \hline
        SDAC-TRPO & 17.559 & 31.248 & 6.1 & 26.343 \\
        SDAC-TRC & 17.277 & 26.341 & 2.58 & 24.618 \\
        \hline 
        RewCor & \bfseries21.78 & 27.241 & 2.78 & 28.429 \\
        CostCor & 18.597 & 22.393 & 2.374 & 25.65 \\
        *\bfseries RewCostCor & 19.336 & \bfseries21.714 & \bfseries1.723 & \bfseries24.581 \\
        \hline
    \end{tabular}
\end{center}
\small{\begin{flushleft}
    \bfseries * proposed
\end{flushleft}}
\caption{PointGoal results for SDAC. The expert algorithm employed for integrating the log-likelihood is indicated alongside SDAC. The abbreviations include RewCoR for cases applying CoR to rewards only, CostCoR for costs only, and RewCostCoR for applying both. SDAC's risk level of CVaR is denoted by SDAC-1.0 and SDAC-0.25.}
\label{CoRselect}
\end{table}

Table \ref{CoRselect} illustrates the outcomes of applying our proposed framework to different components of the system. Implementing the framework exclusively within the reward function yields a positive effect on the overall score but adversely affects the cost sum. Conversely, when the framework is applied solely to the cost function, we observe enhancements in the score, cost sum, and constraint violations (CV). The outcomes derived from employing the log-likelihood probability demonstrate that the exclusive integration of a single type of expert does not exhibit significantly improved performance. When employing the reward expert, SDAC-TRPO, the agent encounters challenges in meeting the constraint threshold, despite exhibiting slightly enhanced performance in reward maximization. However, the utilization of the safe expert, SDAC-TRC, does not demonstrate improved performance in both reward maximization and safety metrics.

The most comprehensive benefits are observed when the CoR term is utilized as outlined in our proposed methodology. This approach results in optimal outcomes for safety metrics, surpassing the results of the other configurations. Although the score marginally decreases compared to its application solely in the reward function, this reduction is a deliberate trade-off to achieve lower constraint violations and cost sums. This strategy underscores the inherent balance between optimizing performance and enhancing safety within safe RL paradigms.

\balance

\section{Conclusion}
In this paper, the safe CoR framework is introduced as an innovative solution to the critical challenges of ensuring safety and reliability in the complex and dynamic environments encountered by autonomous agents. By ingeniously combining reward-focused and safety-oriented expert demonstrations, the safe CoR framework significantly has advanced the field of safe reinforcement learning (safe RL). Our empirical investigations across a variety of environments, including safety gym, metadrive, and the real-world Jackal platform, have demonstrated the framework's remarkable ability to enhance algorithmic performance, while concurrently reducing constraint violations. These results not only underscore the efficacy of the safe CoR framework in balancing performance with safety constraints but also highlight its capability to enhance the domain of autonomous agents and beyond. The contributions of this work, particularly the validation of our framework's superiority in real-world scenarios and its robust applicability across diverse environments, suggest the promising potential for advancements in the development of safe and reliable autonomous agents.

\bibliographystyle{IEEEtran}
\bibliography{myBiB}

\end{document}